\documentclass[final,3p,twocolumn,times,authoryear]{elsarticle}

\usepackage{amssymb}

\usepackage{lineno}
\usepackage{adjustbox}
\usepackage[normalem]{ulem}
\usepackage{hhline}
\usepackage{relsize}
\usepackage{multirow, makecell}
\usepackage{xargs}
\usepackage{algorithmicx}
\usepackage{longtable} 
\usepackage{booktabs}
\usepackage{mathtools,xparse}
\usepackage{color}
\usepackage{subcaption}
\usepackage{url}

\DeclarePairedDelimiter{\vertbfaux}{\bracevert}{\bracevert}

\NewDocumentCommand{\vertbf}{som}{%
	\IfBooleanTF{#1}
	{\vertbfaux*{#3}}
	{\IfNoValueTF{#2}
		{\vertbfaux*{\vphantom{dq}#3}}
		{\vertbfaux[#2]{#3}}%
	}%
}

\newcommand{\mvec}[1]{\mathbf{#1}}

\newcommand{\fracinv}[1]{\frac{1}{#1}}

\newcommand{\tp}{\text{\acrshort{tp}}}
\newcommand{\tn}{\text{\acrshort{tn}}}
\newcommand{\fp}{\text{\acrshort{fp}}}
\newcommand{\fn}{\text{\acrshort{fn}}}
\newcommand{\bvert}{\big\vert}

\newcommand{\acrsmean}[1]{mean \acrshort{#1}}
\newcommand{\acrstotal}[1]{total \acrshort{#1}}
\newcommand{\acrlmean}[1]{mean \acrlong{#1}}
\newcommand{\acrltotal}[1]{total \acrlong{#1}}

\newcommand{\remove}[1]{}

\DeclareMathOperator*{\argmin}{arg\,min}

\newcommandx{\featureCompMean}[4][1=0,2=420]{
	\ifthenelse{\equal{#1}{1}}{\includegraphics[width=#3,viewport={437 #2 860 785}, clip]{#4}}{\includegraphics[width=#3,viewport={0 #2 430 785}, clip]{#4}}
	
}

\usepackage{multirow}

\usepackage{calc}
\usepackage[acronym,nomain]{glossaries}
\newlength\maxlength
\newlength\thislength
\newglossarystyle{mystyle}
{%
	{
		\setlength{\maxlength}{0pt}%
		\forglsentries[\currentglossary]{\thislabel}%
		{%
			\settowidth{\thislength}{\glsentryshort{\thislabel}}%
			\ifdim\thislength>\maxlength
			\setlength{\maxlength}{\thislength}%
			\fi
		}%
		\settowidth{\thislength}{\hspace{1.5em}=\hspace{1em}}%
		\setlength{\glsdescwidth}{\linewidth-\maxlength-\thislength-2\tabcolsep}%
		\begin{longtable}{l@{\hspace{1.5em} \hspace{1em}}p{\glsdescwidth}}
			\toprule
			\multicolumn{1}{l}{\textbf{Abbreviation}} &
			\multicolumn{1}{@{}l}{\textbf{Meaning}}\\%
			\midrule
		}%
		{
			\bottomrule
		\end{longtable}%
	}%
	\renewcommand*{\glsgroupheading}[1]{}%
}
\newacronym{iou}{IoU}{Intersection over Union}
\newacronym{gb_dt}{GBDT}{Gradient Boosted Decision Tree}
\newacronym{ndwi}{NDWI}{Normalized Difference Water Index}
\newacronym{mndwi}{MNDWI}{Modified Normalized Difference Water Index}
\newacronym{lda}{LDA}{Linear Discriminant Analysis}
\newacronym{qda}{QDA}{Quadratic Discriminant Analysis}
\newacronym{sar}{SAR}{Synthetic Aperture Radar}
\newacronym{k_nn}{k-NN}{k Nearest Neighbors}
\newacronym{st}{s.t.}{such that}
\newacronym{iqr}{IQR}{Inter Quartile Range}
\newacronym{sgd}{SGD}{Stochastic Gradient Descent}
\newacronym{ri}{RI}{Rand Index}
\newacronym{nri}{NRI}{Normalized Rand Index}
\newacronym{ncri}{NCRI}{Normalized Class Size Rand Index}
\newacronym{ari}{ARI}{Adjusted Rand Index}
\newacronym{nmi}{NMI}{Normalized Mutual Information}
\newacronym{mi}{MI}{Mutual Information}
\newacronym{ami}{AMI}{Adjusted Mutual Information}
\newacronym{acc}{ACC}{Accuracy}
\newacronym{lsaa}{LSAA}{Linear Sum Assignment \acrlong{acc}}
\newacronym{sse}{SSE}{Sum Of Squared Errors}
\newacronym{tp}{TP}{True Positives}
\newacronym{fp}{FP}{False Positives}
\newacronym{fn}{FN}{False Negatives}
\newacronym{tn}{TN}{True Negatives}
\newacronym{pr}{P}{Precision}
\newacronym{re}{R}{Recall}
\newacronym{f1}{$F_1$}{$F_1$-Score}
\newacronym{hom}{h}{Homogeneity}
\newacronym{com}{c}{Completeness}
\newacronym{v1}{$V_1$}{$V_1$-Measure}
\newacronym{oe}{OE}{Omission Error}
\newacronym{ce}{CE}{Commission Error}
\newacronym{ent}{H}{Entropy}
\newacronym{awei}{AWEI}{Automated Water Extraction Index}
\newacronym{aweish}{AWEISH}{Automated Water Extraction Index for Applications with Shadows}
\newacronym{nir}{NIR}{Near-Infrared}
\newacronym{swir}{SWIR}{Short Wave Infrared}
\newacronym{hsv}{HSV}{Hue Saturation Value}
\newacronym{cndwi}{cNDWI}{Concatenation of \acrshort{ndwi} and \acrshort{mndwi}}
\newacronym{cawei}{cAWEI}{Concatenation of \acrshort{awei} and \acrshort{aweish}}
\newacronym{rgb}{RGB}{Red Green Blue}
\newacronym{api}{API}{Application Programmer Interface}
\newacronym{cma_es}{CMA-ES}{Covariance Matrix Adaption Evolution Strategy}
\newacronym{tpe}{TPE}{Tree-Structured Parzen Estimator}
\newacronym{dem}{DEM}{Digital Elevation Map}
\newacronym{lclu}{LCLU}{Land Cover Land Use}
\newacronym{ndvi}{NDVI}{Normalized Difference Vegetation Index}
\newacronym{svm}{SVM}{Support Vector Machine}
\makeglossaries
\graphicspath{{images/}}

\journal{Remote Sensing of Environment}

\date{February 21, 2023}

\begin{document}
	\begin{frontmatter}
		
		
		
		\title{On the Importance of Feature Representation for Flood Mapping using Classical Machine Learning Approaches}
		
		
		\author[tuk,dfki]{Kevin Iselborn}
		\author[dfki,tuk]{Marco Stricker}
		\author[yamanashi]{Takashi Miyamoto}
		\author[dfki]{Marlon Nuske}
		\author[dfki,tuk]{Andreas Dengel}
		
		\affiliation[tuk]{organization={Department of Computer Science, University of Kaiserslautern-Landau},
			postcode={67663},
			city={Kaiserslautern},
			country={Germany}}
		
		\affiliation[dfki]{organization={German Research Center for Artificial Intelligence (DFKI)},
			postcode={67663}, 
			city={Kaiserslautern},
			country={Germany}}
		
		\affiliation[yamanashi]{organization={Department of Civil and Environmental Engineering, University of Yamanashi},
			postcode={4008511}, 
			city={Kofu},
			country={Japan}}
		
		\begin{abstract}
			Climate change has increased the severity and frequency of weather disasters all around the world so that efforts to aid disaster management activities and recovery operations are of high value. Flood inundation mapping based on earth observation data can help in this context, by providing cheap and accurate maps depicting the area affected by a flood event to emergency-relief units in near-real-time. Modern deep neural network architectures require vast amounts of labeled data for training and whilst a large amount of unlabeled data is available, accurately labeling this data is a time-consuming and expensive task. Building upon the recent development of the Sen1Floods11 dataset, which provides a limited amount of hand-labeled high-quality training data, this paper evaluates the potential of five traditional machine learning approaches such as gradient boosted decision trees, support vector machines or quadratic discriminant analysis for leveraging this data source.
			
			By performing a grid-search-based hyperparameter optimization on 23 feature spaces we can show that all considered classifiers are capable of outperforming the current state-of-the-art neural network-based approaches in terms of \acrstotal{iou} on their best-performing feature spaces, despite our approaches being trained only on the small amount of hand labeled optical and \acrshort{sar} data available in this dataset for performing pixel-wise flood inundation mapping. We show that with total and \acrsmean{iou} values of $0.8751$ and $0.7031$ compared to $0.70$ and $0.5873$ as the previous best-reported results, a simple gradient boosting classifier can significantly improve over the current state-of-the-art neural network based approaches on the Sen1Floods11 test set.
			
			Furthermore, an analysis of the regional distribution of the Sen1Floods11 dataset reveals a problem of spatial imbalance. We show that traditional machine learning models can learn this bias and argue that modified metric evaluations are required to counter artifacts due to spatial imbalance. Lastly, a qualitative analysis shows that this pixel-wise classifier provides highly-precise surface water classifications indicating that a good choice of a feature space and pixel-wise classification can generate high-quality flood maps using optical and \acrshort{sar} data. To facilitate future use of the created feature spaces and the gradient boosting model, we make our code publicly available at: \url{https://github.com/DFKI-Earth-And-Space-Applications/Flood_Mapping_Feature_Space_Importance}
		\end{abstract}
		
		
%
		
%
%
		
	\end{frontmatter}
	
	
	\section{Introduction}
	\label{sec:intro}
	Flood Events are the most frequent and destructive weather disasters (\cite{disastercredHumanCostNatural2015}), which have caused severe human and economic losses over the last years (\cite{huangRapidFloodMapping2020, landuytFloodMappingVegetated2020}), thus causing poverty and destroying lives. Furthermore, due to climate change, these effects will likely get worse in the next years (\cite{wingEstimatesPresentFuture2018}). Recently, flood events in central Europe in the Summer of 2021 have caused severe damage in France, Luxembourg, Belgium, Netherlands, Austria, and Germany (\cite{SaarFRLUXBELGFlood, SWRFloodRP, BelgNetherFlood, OesterreichBayFlood}) and resulted in at least 134 deaths in the Ahrtal, Germany (\cite{swrNochVermissteAktuelle}).
	
	In these situations, quick emergency relief operations are required to aid the local population in need. Flood inundation maps depict the extend of the flood event in a local area. Therefore, accurate and fast to acquire flood maps, are essential to coordinate rescue efforts and are thus of high practical value. In particular, deep learning based flood mapping utilizing satellite imagery is already in use in the Philippines (\cite{delacruzNearRealtimeFloodDetection2020}).
	
	Another application of flood mapping consists of providing information for the development and calibration of hydraulic flood forecasting models (\cite{grimaldiRemoteSensingDerivedWater2016}). Whilst these models have shown potential for risk assessment (\cite{wingEstimatesPresentFuture2018}) and many solutions have been proposed for simulating flood extend (\cite{merwadeIntegratedApproachFlood2017, yamazakiPhysicallyBasedDescription2011, liApplicationRemoteSensing2016}), verification of these models remains challenging as the number of global gauge stations keeps decreasing (\cite{grimaldiRemoteSensingDerivedWater2016}). In this context, flood mapping can not only help by mitigating the effects of flood events but also by supporting the forecasting of flood effects.
	
	As a data source for flood mapping, remote sensing for earth-observation has shown promising results (\cite{shenInundationExtentMapping2019, tsyganskayaSARbasedDetectionFlooded2018, grimaldiRemoteSensingDerivedWater2016}). Remote sensing hereby refers to the process of measuring physical characteristics of an area at a distance (\cite{WhatRemoteSensing}), which for the purpose of flood mapping consists of observing the earth-surface using sensors on aerial vehicles or on satellites.
	
	In the context of flood mapping, aerial imagery is typically considered the most reliable source of flood extend data, however, the requirement to perform additional flights for data acquisition may make this impractical, especially in developing countries (\cite{grimaldiRemoteSensingDerivedWater2016}). In contrast, satellite data is available at high velocity and volume in the form of continuous global coverage of the earth-surface. For example, free of charge world-wide satellite data is provided by the European Space Agency's Copernicus Program (\cite{esaCopernicusCopernicus}). Furthermore, different sensors such as \acrlong{sar} (\cite{esaSentinel1MissionsInstrument}) or multispectral instruments (\cite{esaSentinel2UserGuide}) are available in varying resolutions (\cite{planetscopeSatelliteImageryArchive2021, esaSentinel2UserGuide, LandsatMssions}) thereby providing a large variety of data.
	
	The availability of these three of the four V's of big data (volume, variety, velocity, veracity) (\cite{diazDiscoverBigData2020}), missing only veracity, results in a need for highly-scalable and cheap solutions requiring no-human interaction. Artificial Intelligence has been shown to provide solutions to these goals in many different application scenarios (\cite{ImageNet, PASCALVOC, MSCOCO}), exceeding human performance (\cite{yuCoCaContrastiveCaptioners2022}).
	
	Traditionally, these approaches are based on thresholding algorithms, region growing or multi-temporal change detection (\cite{shenInundationExtentMapping2019}), however more recently Machine Learning approaches such as Random Forests, Support Vector Machines or Deep Learning are also entering the scene (\cite{tsyganskayaSARbasedDetectionFlooded2018, delacruzNearRealtimeFloodDetection2020, konapalaExploringSentinel1Sentinel22021, bentivoglioDeepLearningMethods2021, baiEnhancementDetectingPermanent2021}). 
	
	Whilst historical flood databases exist, for example provided by the Dartmouth Flood Observatory (\cite{universityofcoloradoFloodObservatory}), these are based on low-quality automatic labels (\cite{grimaldiRemoteSensingDerivedWater2016}) and machine learning ready datasets were missing up until recently (\cite{bonafiliaSen1Floods11GeoreferencedDataset2020, rambourFloodDetectionTime2020, mateo-garciaGlobalFloodMapping2021, drakonakisOmbriaNetSupervisedFlood2022}).
	
	This has resulted in many studies operating on data that was specifically acquired for the individual paper (\cite{tsyganskayaSARbasedDetectionFlooded2018, bentivoglioDeepLearningMethods2021}), furthermore, due to the time and effort required for manually creating flood segmentation maps, accurate ground truth data is rare. Some studies try to avoid this time effort by both training their models and computing their evaluation metrics with respect to automatically generated labels (\cite{delacruzNearRealtimeFloodDetection2020, huangRapidFloodMapping2020}). 
	
	In 2020, Bonafilia et al. proposed the Sen1Floods11 dataset, which contains a limited amount of high-quality hand-labeled data, in combination with a larger weakly labeled dataset (\cite{bonafiliaSen1Floods11GeoreferencedDataset2020}). Whilst the labels in this subset are noisy, due to being created with ``weak'' thresholding approaches, training with inaccurate (noisy) supervision (\cite{zhouBriefIntroductionWeakly2018a}) has been shown to significantly improve the performance of flood mapping methodologies (\cite{baiEnhancementDetectingPermanent2021}). Moreover, Sen1Floods11 offers data originating from flood events all around the world and spanning all major biomes, which is in contrast to many previously used data sources that consider only a single region (\cite{tsyganskayaSARbasedDetectionFlooded2018, bentivoglioDeepLearningMethods2021}) or datasets that focus only on a subset of the worlds biomes (\cite{rambourFloodDetectionTime2020, mateo-garciaGlobalFloodMapping2021, drakonakisOmbriaNetSupervisedFlood2022}).  
	
	So far there are few publications utilizing this dataset (\cite{patelEvaluatingSelfSemiSupervised2021, baiEnhancementDetectingPermanent2021, konapalaExploringSentinel1Sentinel22021, katiyarNearRealTimeFloodMapping2021, yadavAttentiveDualStream2022, jainMultimodalContrastiveLearning2022}). However, the access to hand-labeled data from flood events spanning all major biomes (\cite{bonafiliaSen1Floods11GeoreferencedDataset2020}) offers a unique opportunity to test and evaluate the performance of flood mapping machine learning models on a global scale. 
	
	Based on this dataset, this paper explores the potential of traditional machine learning algorithms to leverage this limited amount of data for performing flood mapping, thereby providing a fast-to-execute flood-mapping methodology as well as a new state-of-the art method on the Sen1Floods11 dataset using a gradient boosting classifier based on combined optical and \acrshort{sar} data. Quantitative results of all tested classifiers are provided and can serve as baselines for future work. An extensive evaluation further highlights the benefits of the best performing methodology by showing that this pixelwise classifier is capable of computing highly precise surface water segmentation maps. 
	
	\section{Related Work}
	\label{sec:related}
	We will now provide an overview over related work in the area of flood inudation mapping with a focus on the Sen1Floods11 dataset. Section \ref{sec:related:classical} briefly reviews classical techniques for flood inundation mapping and Section \ref{sec:related:ml} then summarizes deep learning based approaches that have been applied to perform flood inundation mapping on the Sen1Floods11 dataset. Lastly, we review other work that has been carried out towards using feature spaces for flood inundation mapping in Section \ref{sec:related:features}.
	\subsection{Classical Methodologies} \label{sec:related:classical}
	Water can often be identified in \acrshort{sar} images by a low electromagnetic reflectance value. This has led to the development of many approaches based on a threshold value that classifies individual pixels as water or non-water.
	
	Whilst the determination of this threshold is done manually in some cases (\cite{martinisOperationalRealtimeFlood2009,chiniHierarchicalSplitBasedApproach2017, tsyganskayaSARbasedDetectionFlooded2018, RegionGrowingAndChangeDetection, shenInundationExtentMapping2019}), it is more often automated by using a threshold determination method (\cite{tsyganskayaSARbasedDetectionFlooded2018}) such as the one developed by \cite{otsuThresholdSelectionMethod1979} or that of \cite{kittlerMinimumErrorThresholding1986}. For automated thresholding approaches such as Otsu's method (\cite{otsuThresholdSelectionMethod1979}), a sufficiently bimodal image histogram is necessary to extract a useful threshold value. 
	
	However, in practice often large overlapping value ranges for the flood water and dry land classes can be observed so that no sufficiently bimodal image histogram can be computed (\cite{tsyganskayaSARbasedDetectionFlooded2018}). A common approach to alleviate this effect is to subdivide the region of interest into a suitable set of tiles and then calculate the threshold on a selected subset of these (\cite{martinisOperationalRealtimeFlood2009, shenInundationExtentMapping2019}). Furthermore, different geographical regions might require largely different threshold values due to differences in the backscatter intensity to achieve a sufficient segmentation (\cite{bonafiliaSen1Floods11GeoreferencedDataset2020}), further hampering transferability of these simple approaches.
	\subsection{Deep Learning for Flood Mapping}\label{sec:related:ml}
	In more recent years, deep learning has been entering the flood mapping scene (\cite{bentivoglioDeepLearningMethods2021}). Deep Neural Networks can learn appropriate feature representations for the given classification task at hand (\cite{bentivoglioDeepLearningMethods2021, patelEvaluatingSelfSemiSupervised2021}), hence overcoming problems caused for example by insufficiently bimodal histograms or variations in the backscatter intensity, if a sufficiently large amount of data is provided. 
	
	As accurately labeling large amounts of satellite imagery is time-consuming, there have been attempts to leverage automatically generated weak labels for training neural network based approaches (\cite{bonafiliaSen1Floods11GeoreferencedDataset2020, huangRapidFloodMapping2020, baiEnhancementDetectingPermanent2021}). In particular, \cite{bonafiliaSen1Floods11GeoreferencedDataset2020} provide results for a Resnet-50 baseline model in the paper describing the dataset used in our study, showing that a simple setup using an off-the-shelf architecture can outperform traditional \acrshort{sar} thresholding methods. 
	
	\cite{katiyarNearRealTimeFloodMapping2021} improve upon this by training U-Net (\cite{UNet}) on the weakly labeled dataset provided by Sen1Floods11 and finetuning it to the hand-labeled split. \cite{baiEnhancementDetectingPermanent2021} further enhance this by using BASNet, a U-Net style image segmentation network (\cite{qinBASNetBoundaryAwareSalient2019}), in combination with a hybrid loss function, consisting of the structural similarity loss, \acrshort{iou} loss, and focal loss. To the best of our knowledge, the reported \acrsmean{iou} of $0.5873$ (\cite{baiEnhancementDetectingPermanent2021}) forms the best \acrsmean{iou} value for supervised learning on the Sen1Floods11 dataset, which the \acrfull{gb_dt} model in Section \ref{sec:classifiers:models} is able to exceed with a relative improvement of $19.7\%$. 
	
	With the advent of contrastive learning based deep learning (\cite{zhongDeepRobustClustering2020, huangDeepSemanticClustering2020, patelEvaluatingSelfSemiSupervised2021, jungContrastiveSelfSupervisedLearning2022}), \cite{patelEvaluatingSelfSemiSupervised2021} explore unsupervised flood segmentation based on \acrshort{sar} images. In particular, they explore the usage of self-supervised and semi-supervised learning based on SimCLR (\cite{chenSimpleFrameworkContrastive2020}) and FixMatch (\cite{sohnFixMatchSimplifyingSemiSupervised2020}) for riverbed segmentation, land cover land use classification, and flood inundation mapping on the Sen1Floods11 dataset (\cite{patelEvaluatingSelfSemiSupervised2021}). For this DeepLabv3+ (\cite{chenEncoderDecoderAtrousSeparable2018}) is first pre-trained on a large dataset of either \acrshort{sar} or optical images using self-supervised training, before being finetuned to the downstream task. \cite{jainMultimodalContrastiveLearning2022} expand on this procedure by using different data modalities sampled at the same spatial location as positive pairs for contrastive learning, before finetuning to the downstream task. 
	
	\cite{yadavAttentiveDualStream2022} use an attentive dual stream siamese network to perform \acrshort{sar} change detection based flood inundation mapping. For this additional pre-flood \acrshort{sar} imagery for the locations present in the Sen1Floods11 dataset were acquired to achieve a \acrstotal{iou} of $0.70$, which is exceeded by our \acrshort{gb_dt} model with a relative improvement of $25.01\%$, despite it using fewer data-sources by operating only on post-flood images. Whilst \cite{yadavAttentiveDualStream2022} do not provide a \acrsmean{iou} value, they exceed the \acrstotal{iou} of $0.6452$ reported in the work by \cite{baiEnhancementDetectingPermanent2021}, so that we consider this the current state-of-the-art result on the given dataset.
	
	\subsection{Feature Spaces for Flood Mapping using Machine Learning Approaches}\label{sec:related:features}
	While all machine learning approaches mentioned so far have focused on training neural networks to create feature extractors based on raw satellite data, \cite{konapalaExploringSentinel1Sentinel22021} provide the neural network with already extracted features based on \acrshort{sar}, optical and DEM data. Specifically, they show that domain knowledge in the form of simple transformations of optical data can increase the performance of the model. Unfortunately, these results cannot be compared to other works on the Sen1Floods11 dataset, since \cite{konapalaExploringSentinel1Sentinel22021} use a different evaluation procedure based on splits which differ significantly from those provided by \cite{bonafiliaSen1Floods11GeoreferencedDataset2020}.
	
	Machine Learning approaches such as \acrfull{svm} and Random Forest have also been applied to the problem of flood inundation mapping (\cite{tsyganskayaSARbasedDetectionFlooded2018, huangRapidFloodMapping2020}) in some cases leveraging domain knowledge in the form of a few hand-crafted feature spaces (\cite{manakosFusionSentinel1Data2020, slagterMappingWetlandCharacteristics2020}). Providing classical machine learning algorithms with hand-crafted features has had a long tradition in the computer vision community and these algorithms are known to perform well when given high-quality features as input, whilst requiring much less training data than a deep neural network. This raises the question, whether a machine learning algorithm trained using domain-knowledge in the form of feature spaces similar to those used by \cite{konapalaExploringSentinel1Sentinel22021} can achieve similiar or better performance on the small amount of high-quality training data available in the Sen1Floods11 dataset. Furthermore, comparable studies of classical machine learning methods using a single benchmarking dataset are missing so far (\cite{tsyganskayaSARbasedDetectionFlooded2018}).
	
	Our work therefore provides three main contributions. First of all, we provide quantitative results for five classical machine learning approaches using eight metrics based on an extensive grid-search-based hyperparameter optimization to find both the best performing feature spaces and their corresponding hyperparameters for all data modalities available in the Sen1Floods11 dataset (\acrshort{sar}, optical and combined \acrshort{sar} and optical data). Secondly, we provide an extensive analysis of the best performing model to analyze the properties of machine learning approaches trained on remote sensing feature spaces. We show that the tested machine learning algorithms outperform deep learning models trained using additional data-sources, such as pre-flood images, or additional training data. Lastly, in the course of our analysis, we show that the spatial imbalance present in the Sen1Floods11 dataset can result in a bias both in the resulting classifier and the evaluated metrics. We propose approaches for evaluating the metrics in an unbiased manner, that can also be employed by future work. The considered flood mapping methodologies exceed the existing state of the art approaches and can thus either be deployed as highly-scalable classifiers or can be used to train deep learning approaches with high-quality labeled data.
	
	\section{Data}
	\label{sec:data}
	Throughout this work, we use two types of earth-observation data, which we describe in this Section. Optical data is described in Section \ref{sec:data:optical_data} and \acrshort{sar} data is described in Section \ref{sec:data:sar}. Lastly, we describe the dataset used throughout this work in Section \ref{sec:sen1floods11}.
	
	\subsection{Optical Data}\label{sec:data:optical_data}
	\begin{table*}
		\centering
		\caption[Available observation channels and their spatial resolution of the Sentinel-2 satellites]{Available 13 observation channels and their spatial resolution of the Sentinel-2 satellite constellation (\cite{esaSentinel2UserGuide}). Notice that Sentinel-2 satellite constellation features two satellites, however as the bands differ only insignificantly from each other, only Sentinel-2A is depicted here.}
		\resizebox{0.75\textwidth}{!}{
			\begin{tabular}{ccccc}
				\toprule
				\textbf{Band Number} & \textbf{Band Name} & \textbf{Central wavelength (in nm)} & \textbf{Bandwidth (in nm)} & \textbf{Spatial Resolution (in m)} \\\midrule
				1 & Coastal & 442.7 & 21 & 60 \\
				2 & Blue & 492.4 & 66 & 10 \\
				3 & Green & 559.8 & 36 & 10 \\
				4 & Red & 664.6 & 31 & 10 \\
				5 & RedEdge-1 & 704.1 & 15 & 20 \\
				6 & RedEdge-2 & 740.5 & 15 & 20 \\
				7 & RedEdge-3 & 782.8 & 20 & 20 \\
				8 & \acrshort{nir} & 832.8 & 106 & 10 \\
				8a & Narrow \acrshort{nir} & 864.7 & 21 & 20 \\
				9 & Water Vapor & 945.1 & 20 & 60 \\
				10 & Cirrus & 1373.5 & 31 & 60 \\
				11 & \acrshort{swir}-1 & 1613.7 & 91 & 20 \\
				12 & \acrshort{swir}-2 & 2202.4 & 175 & 20 \\\bottomrule
			\end{tabular}
		}
		\label{tab:intro:optical_data:s2_bands}
	\end{table*}
	The most straightforward data source for earth-observation imagery is optical data. Many satellites provide \acrshort{rgb}-data that is easy-to-interpret by humans, often allowing the separation of large water bodies at a glance. However, many modern satellites provide multi-spectral instruments (\cite{esaSentinel2UserGuide, LandsatMssions}), giving access to measurements of light-wavelengths outside the human visible spectrum. Leveraging the information contained in these channels is more challenging for humans, but by employing suitable feature transformations machine learning algorithms can take advantage of them, as is shown by the results presented in Section \ref{sec:classifiers:quantitative}. 
	
	The Sentinel-2 optical data provides thirteen channels with wavelengths ranging from the near ultra-violet ($\approx 442.7nm$) into the short wave infrared spectrum ($\approx 2202.4nm$) as depicted in Table \ref{tab:intro:optical_data:s2_bands}.
	Each pixel on the ground represents a square ranging from $\approx 10\times 10$m  to $\approx 60\times 60$m per pixel. Only the Red, Green, Blue, and \acrlong{nir} bands achieve the full resolution (\cite{esaSentinel2UserGuide, esaSentinel2TechnicalGuide}). 
	
	Flood mapping based on optical data has been shown in the literature (\cite{konapalaExploringSentinel1Sentinel22021,baiEnhancementDetectingPermanent2021}) to be extremely effective, as long as clear sight of the earth's surface can be provided. As flood events are often accompanied by cloudy conditions, flood mapping approaches based on optical data are less applicable for emergency relief operations. However, the high-quality flood maps that can be generated after the weather has cleared, are of high importance for calibrating hydraulic simulations or for creating label data to train deep learning algorithms. 
	\subsection{\acrfull{sar} Data}\label{sec:data:sar}
	\begin{figure*}
		\centering
		\begin{subfigure}{0.45\linewidth}
			\centering
			\includegraphics[width=0.45\linewidth]{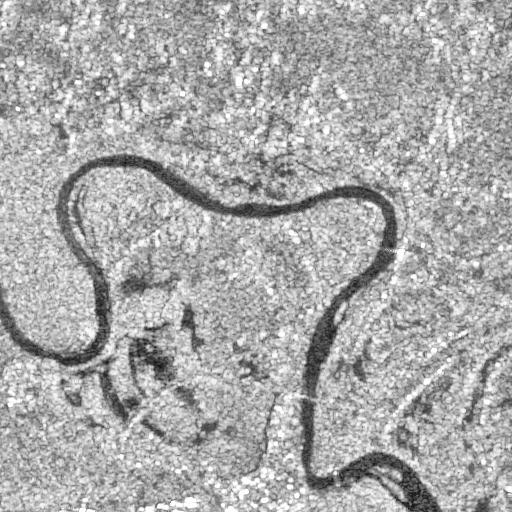}
			\includegraphics[width=0.45\linewidth]{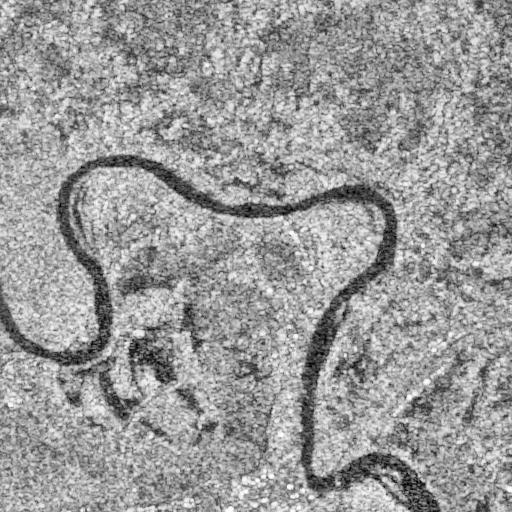}
			\caption{Exemplary Sentinel-1 (\acrshort{sar}) VV and VH polarization}
		\end{subfigure}
		\begin{subfigure}{0.45\linewidth}
			\centering
			\includegraphics[width=0.45\linewidth]{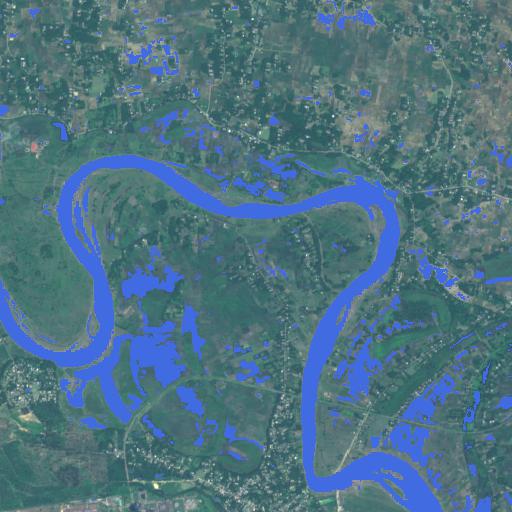}
			\caption{Ground truth water labels depicted on a Sentinel-2 RGB composite}
		\end{subfigure}
		\caption[Exemplary Sentinel-1 \acrshort{sar} and Sentinel-2 RGB images from the Sen1Floods11 dataset]{Exemplary \acrshort{sar} image and ground truth water labels from a flood location in India present in the Sen1Floods11 dataset \cite{bonafiliaSen1Floods11GeoreferencedDataset2020}. It is observed, that the \acrshort{sar} images highlight most of the open water present in the images, due to the lower backscatter intensity of water. }
		\label{fig:eo_data:exemplary_sar_opt}
	\end{figure*}
	A different data source is \acrfull{sar} based satellite data. \acrshort{sar} sensors, such as those provided by the Sentinel-1 satellite, operate by measuring the reflectance of actively sent electromagnetic waves (\cite{esaSentinel1MissionsInstrument}) often called backscatter (\cite{shenInundationExtentMapping2019}). A dual polarization \acrshort{sar}-Sensor such as Sentinel-1 provides two data channels in the form of VV (Vertical-Vertical) and VH (Vertical-Horizontal) polarization that can be employed for flood inundation mapping. \acrshort{sar}-sensors are capable of observing the ground even under cloudy conditions and during nighttime, resulting in \acrshort{sar}-sensors being a preferable data source for near-real-time applications.
	
	As can be observed in Figure \ref{fig:eo_data:exemplary_sar_opt}, water generally provides lower backscatter values, so that regions covered by surface water are easy to distinguish from surrounding objects. This property has inspired simple but effective classical methodologies such as the \acrshort{sar} thresholding described in Section \ref{sec:related:classical} (\cite{shenInundationExtentMapping2019, tsyganskayaSARbasedDetectionFlooded2018}).
	
	However, \acrshort{sar} data also contains some error sources uncommon to traditional \acrshort{rgb} optical data. For example interactions between reflected electromagnetic waves from scattering objects in neighboring pixels can result in noise like speckles. These interferences, which can also be observed to some extent in Figure \ref{fig:eo_data:exemplary_sar_opt}, are especially problematic for pixel-wise algorithms that cannot take larger structures in the image into account.
	
	To correct for speckle noise, many filtering algorithms have been developed (\cite{leeSpeckleAnalysisSmoothing1981, leeRefinedFilteringImage1981, kuanAdaptiveRestorationImages1987, vasileIntensitydrivenAdaptiveneighborhoodTechnique2006, leeImprovedSigmaFilter2009, banerjeeComprehensiveSurveyFrost2020}). We use the improved lee sigma filter (\cite{leeImprovedSigmaFilter2009}) that has for example been used by \cite{landuytFloodMappingVegetated2020}.
	
	\subsection{The Sen1Floods11 Dataset}\label{sec:sen1floods11}
	\cite{bonafiliaSen1Floods11GeoreferencedDataset2020} developed Sen1Floods11 to serve as a benchmark dataset for flood mapping algorithms that are capable of performing well all around the world. To achieve this, the authors selected eleven flood events from the Dartmouth flood observatory database (\cite{DarmouthFloodDB}) such that all major biomes are covered by the acquired satellite imagery. For these flood events, \cite{bonafiliaSen1Floods11GeoreferencedDataset2020} then collected  Sentinel-1 \acrshort{sar} and Sentinel-2 optical post-flood data, keeping only the intersections of the stacked and georeferenced (WGS 84 projection) imagery (\cite{bonafiliaSen1Floods11GeoreferencedDataset2020}). As mentioned in Section \ref{sec:data:optical_data}, the optical data is only available in differing spatial resolutions. In order to provide all data in a machine learning ready form, bands with a resolution lower than $10\times 10$m were upsampled by \cite{bonafiliaSen1Floods11GeoreferencedDataset2020}, before including them in the Sen1Floods11 dataset.
	
	From this data, regions mainly affected by flooding and with minority cloud cover were selected by remote sensing analysts. These were then divided into $4,831$ $512\times 512$ images. Out of this resulting dataset, $446$ were hand-labeled by remote sensing analysts as either surface water (including both flood and permanent water), dry land, or no-data. Hereby no data pixels include cloud masks, pixels masked out due to lack of either Sentinel-1 or Sentinel-2 data or pixels that could not be confidently identified by the labeling analyst as either inundated or dry. 
	
	\subsubsection{Value distribution} \label{sec:sen1floods11:value_dist}
	\begin{figure}
		\centering
		\includegraphics[width=0.8\linewidth]{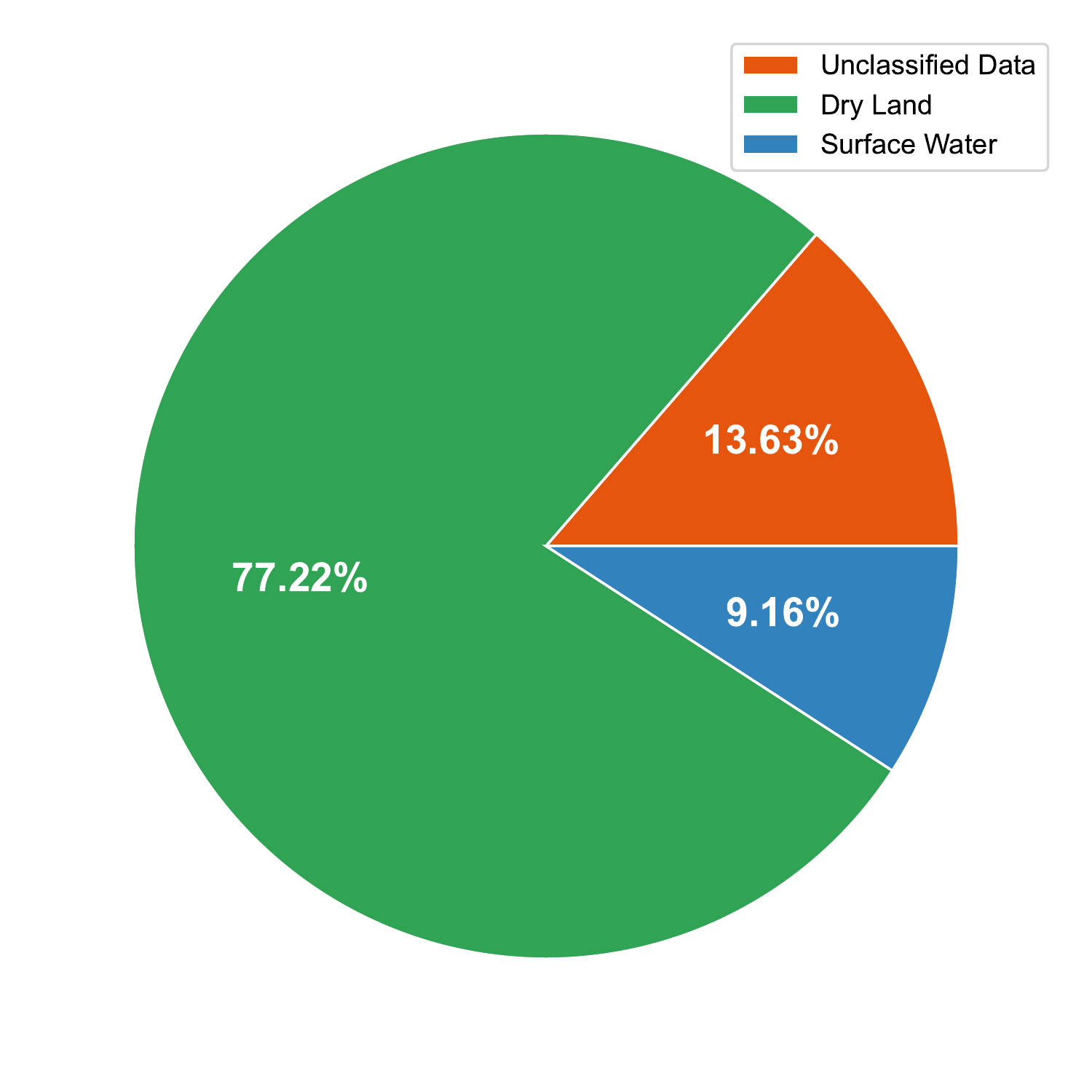}
		\caption{The percentage of total pixels that are marked as dry ($77.22\%$) or water ($9.18\%$) or which could not be classified by the labeling remote sensing analysts ($13.63\%$).}
		\label{fig:sen1floods11_value_dist:flood_pixel_dist}
	\end{figure}
	The first question to answer, given this high-quality labeled data, is how the distribution of the Sen1Floods11 dataset looks like. As depicted in Figure \ref{fig:sen1floods11_value_dist:flood_pixel_dist}, only a small fraction of the hand-labeled dataset consists of flooded pixels. 
	
	This tail distribution is problematic, as a classifier trained on such a dataset will tend towards learning solutions, that prefer classifying something as dry land instead of as flooded. However, the resulting low \acrlong{re} classifiers are undesirable, especially in the context of emergency-relief operations, as omitting certain regions from the flood inundation map may result in life losses. 
	
	Therefore strategies, which are mainly based on re-weighting the individual classes within the optimization objective of the given algorithm, have been proposed in flood mapping literature and shown their success at improving the mapping performance (\cite{chiniHierarchicalSplitBasedApproach2017, shaerikarimiApplicationMachineLearning2019, baiEnhancementDetectingPermanent2021}). However, as we will observe in Section \ref{sec:classifiers:models}, for our tested classical machine learning approaches, this is not always beneficial.
	\subsubsection{Spatial Distribution}\label{sec:intro:sen1floods11:spatial_dist}
	\begin{figure}
		\includegraphics[height=0.3\textheight]{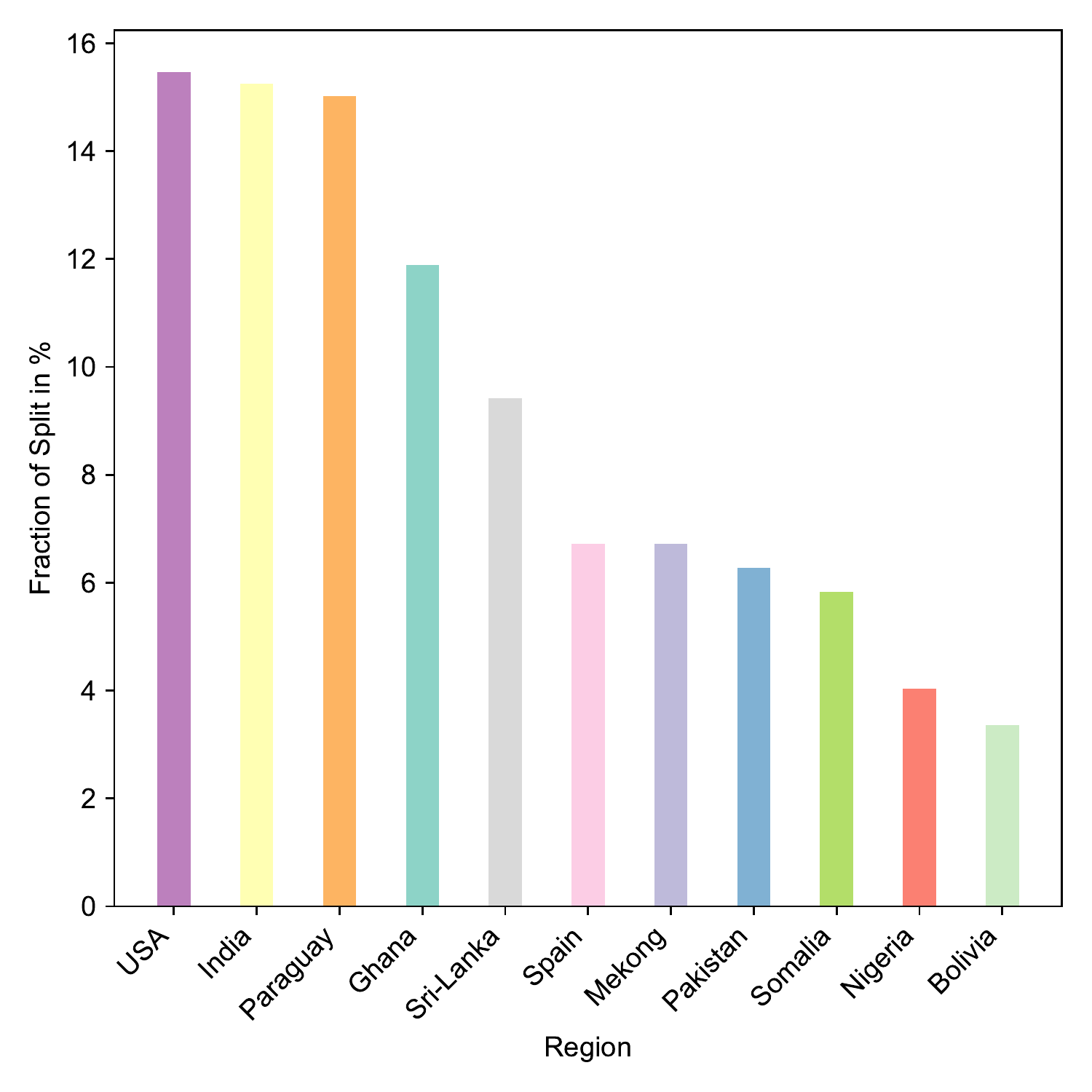} 
		\caption{Spatial distribution of the Sen1Floods11 dataset per-country depicted as the relative area of each region of the complete hand-labeled Sen1Floods11 dataset.}
		\label{fig:intro:sen1floods11_spatial_dist:region_dist}
	\end{figure}
	
	Further insights can be gained by inspecting statistics for the individual regions in the Sen1Floods11 dataset. In Figure \ref{fig:intro:sen1floods11_spatial_dist:region_dist}, a strong region-imbalance can be observed. Whilst images from the USA take up almost $16\%$ of the hand-labeled dataset, only very few images originating from Nigeria are present. 
	
	This can have an effect on the calculation of metrics, where a good performance of a flood mapping model on the majority regions, can result in overestimation of the true performance of the model in a global context. Moreover, algorithms trained on this dataset with the given bias, may learn to neglect minority areas. Thus constructing flood mapping algorithms on this dataset, that are agnostic to the region they operate on, is hard to achieve. We therefore specifically analyze our model using a novel metric calculation procedure, to ensure a good performance of the proposed classifier all around the world. To the best of our knowledge such an analysis has not been carried out for the Sen1Floods11 dataset, even though as is shown in Section \ref{sec:classifiers:quantitative}, models such as the proposed \acrshort{gb_dt}-classifier can learn this bias.
	
	\section{Methodology} \label{sec:classifiers}
	To analyze the performance of machine learning models on the Sen1Floods11 dataset, in the following Section \ref{sec:classifiers:metrics} we explain our evaluation metrics and experiment setup. We then describe the features and their combinations, referred to as feature spaces, which have been explored in this work in Section \ref{sec:data:feature_spaces}. Lastly, we describe the evaluated models and their hyperparameter optimization in Section \ref{sec:classifiers:models}, before performing a quantitative and qualitative evaluation of the tested algorithms on the described feature spaces in the Section \ref{sec:results}.

	\subsection{Evaluation Methodology}
	\label{sec:classifiers:metrics}
	Throughout our experiments we use the pre-defined 60:20:20 train-validation-test splits of the hand-labeled data provided by the authors of the Sen1Floods11 dataset. Models are trained solely using the hand-labeled train data, with further grid-search based hyperparameter optimization using metrics as calculated on the validation split and averaged over four seeds, to avoid biasing the test set results (\cite{deisenrothMathematicsMachineLearning2020}). This helps to provide a fair comparison of the classical machine-learning algorithms with state-of-the-art neural network based approaches, despite these having a much larger model capacity. 
	
	In addition to comparative plots generated from these validation set results, we provide final evaluations using 16 seeds in tabular format for each of the different data-modalities that are available on the Sen1Floods11 dataset (\acrshort{sar}, optical and combined data) and present for each of these modalities the model according to the highest validation-set \acrsmean{iou}. 
	
	Results on the additional Bolivia test split in the Sen1Floods11 dataset, containing data from a region that was not observed during training (\cite{bonafiliaSen1Floods11GeoreferencedDataset2020}), are also provided in order to facilitate estimating the performance of the given algorithm on a new geographical location. Cross-validation procedures are not used to save processing time and make the tuning process more comparable to that of other past and future works utilizing the pre-defined splits.
	
	\subsubsection{Metrics}
	A commonly used metric for the evaluation of classifiers that intuitively represents the fraction of errors made by the classifier is \acrfull{acc} (\cite{bonafiliaSen1Floods11GeoreferencedDataset2020}), which is defined in terms of \acrfull{tp}, \acrfull{tn}, \acrfull{fp} and \acrfull{fn} for the surface water class in equation \ref{eq:evaluation:accuracy}.
	\begin{equation}
		\text{\acrshort{acc}} =\frac{\tp + \tn}{\tp + \fp + \tn + \fn} \label{eq:evaluation:accuracy}
	\end{equation}
	However, as pointed out by the authors of Sen1Floods11, Accuracy is not a good estimator of the performance of flood mapping classifiers (\cite{bonafiliaSen1Floods11GeoreferencedDataset2020}) due to the class imbalance problem mentioned in Section \ref{sec:sen1floods11:value_dist}. Substituting a classifier that predicts everything as dry land and the class distribution of Sen1Floods11 into equation \ref{eq:evaluation:accuracy} would still achieve an accuracy value of $\approx 89.4\%$.
	
	In order to overcome this problem, \cite{bonafiliaSen1Floods11GeoreferencedDataset2020} propose to use \emph{\acrfull{iou}} as the main metric for the Sen1Floods11 dataset. 
	\begin{equation} 
		\text{\acrshort{iou}} = \frac{\tp}{\tp + \fp + \fn} \label{eq:evaluation:iou}
	\end{equation}
	Intuitively \acrshort{iou} describes the percentage of the area that is correctly identified as inundated (the intersection) out of the area that is actually water or has incorrectly been recognized as water (the union). Therefore it is unaffected by large portions of the images containing only dry land. 
	
	\cite{konapalaExploringSentinel1Sentinel22021} base their evaluation on \emph{\acrfull{pr}}, \emph{\acrfull{re}}, and \emph{\acrfull{f1}}. Recall hereby measures the percentage of inundated pixels that were recognized by the flood mapping algorithm, whereas Precision measures the fraction of pixels that have been correctly identified as water, out of all surface water classifications. 
	\begin{align}
		\text{\acrshort{pr}} &= \frac{\tp}{\tp + \fp} \\
		\text{\acrshort{re}}  &= \frac{\tp}{\tp + \fn} \\
		\text{\acrshort{f1}} &= \frac{2 \cdot \text{\acrshort{pr}} \cdot \text{\acrshort{re}}}{\text{\acrshort{pr}} + \text{\acrshort{re}}} = \frac{2\cdot \tp}{2 \cdot \tp + \fp + \fn} \label{eq:classifiers:metrics:formulas:f_score}
	\end{align}
	
	In addition to \acrshort{iou}, both \cite{bonafiliaSen1Floods11GeoreferencedDataset2020} and \cite{baiEnhancementDetectingPermanent2021} also provide the \emph{\acrfull{oe}} and \emph{\acrfull{ce}} rates for the evaluated methods. These metrics represent how many pixels were incorrectly not recognized as inundated or dry land respectively. For these we define $\tp_{\text{Dry}}, \fn_{\text{Dry}}$ as the \acrlong{tp} and \acrlong{fn} rates for the dry land class, corresponding to $\tn$ and $\fp$ of the surface water class.
	\begin{align}
		\text{\acrshort{oe}} &:= \frac{\fn}{\fn + \tp} = \frac{\fn + \tp - \tp}{\fn + \tp} = 1 - \text{\acrshort{re}} \label{eq:classifiers:metrics:formulas:omission}\\
		\text{\acrshort{ce}}  &:= \frac{\fp}{\fp + \tn} = \frac{\fn_{\text{Dry}}}{\fn_{\text{Dry}} + \tp_{\text{Dry}}} \stackrel{\ref{eq:classifiers:metrics:formulas:omission}}{=} 1 - \text{\acrshort{re}}_{\text{Dry}}\label{eq:classifiers:metrics:formulas:commission}
	\end{align} 
	
	It can be observed in equations \ref{eq:classifiers:metrics:formulas:omission} and \ref{eq:classifiers:metrics:formulas:commission} that the Ommission and \acrlong{ce} rates are equivalent to $1-$ the \acrlong{re} of the flooded or dry land classes, respectively. Therefore, we only provide the equivalent \acrlong{re} values instead of the Ommission and \acrlong{ce} rates, so that all metrics reported in this work consistently indicate better performance via larger values.
	
	\subsubsection{Reported Metrics and Metric Aggregation}\label{sec:classifiers:metrics:formulas:eval}
	It has to be noted that different ways of calculating the individual metrics are possible. Measures can be calculated for each of the images individually and then aggregated across the evaluation dataset, yielding a mean and standard deviation of the measure. Alternatively, classifiers can first aggregate errors from all images in the evaluation split by summing up the \acrshort{tp}, \acrshort{tn}, \acrshort{fp} and \acrshort{fn} values and evaluating the individual metric on the result, yielding just a single value. We will refer to metrics of the first kind as \emph{mean-based metrics} and depict them for example as \acrsmean{iou}, whereas metrics calculated using the second procedure will be called \emph{total metrics} such as \acrstotal{iou}.
	
	\acrlong{acc} and \acrshort{iou} are reported both as mean-based and total metrics, whilst \acrlong{f1}, \acrlong{re}, \acrlong{pr} of the surface water class and the \acrlong{re} of the dry land class are reported only as total metrics. This provides us with insights into the stability of the algorithms performance across the evaluation split using the mean and standard deviation of the mean-based measures whilst also providing full error estimates using the total metrics as well as permitting comparisons with previous work carried out on the Sen1Floods11 dataset. 
	
	\subsection{Tested Feature Spaces} \label{sec:data:feature_spaces}
	To improve the quality of surface and flood water detection, many features based on optical data have been developed and used in the literature (\cite{mcfeetersUseNormalizedDifference1996, xuModificationNormalisedDifference2006, feyisaAutomatedWaterExtraction2014}). \cite{konapalaExploringSentinel1Sentinel22021} show that flood mapping approaches based on hand-crafted feature spaces can outperform those based on raw optical or \acrshort{sar} data (\cite{konapalaExploringSentinel1Sentinel22021}). This Section, therefore, describes the used feature spaces and especially the transforms performed on optical satellite data.
	
	A common method of analysis for remote sensing imagery based on multi-spectral instruments is the analysis of a corresponding remote sensing \emph{index} (\cite{hueteVegetationIndicesRemote2012}). These indexes form simple and straight forward to implement functions of subsets of the available multi-spectral bands. For the detection of water, the \emph{\acrfull{ndwi}} has been proposed (\cite{mcfeetersUseNormalizedDifference1996}). This index is based upon surface water having a lower reflection in the \acrshort{nir} band, whilst at the same time producing a high reflection in the green bands (\cite{mcfeetersUseNormalizedDifference1996}).
	\begin{equation}
		\text{\acrshort{ndwi}} = \frac{\text{GREEN} - \text{\acrshort{nir}}}{\text{GREEN} + \text{\acrshort{nir}}}
	\end{equation}
	As remarked by \cite{xuModificationNormalisedDifference2006}, the \acrshort{ndwi} is susceptible to overprediction in urban areas. By using the \acrshort{swir} band, instead of the \acrshort{nir} one, this effect can be reduced (\cite{xuModificationNormalisedDifference2006, konapalaExploringSentinel1Sentinel22021}).
	
	\begin{equation}
		\text{\acrshort{mndwi}} = \frac{\text{GREEN} - \text{\acrshort{swir}-1}}{\text{GREEN} + \text{\acrshort{swir}-1}}
	\end{equation}
	
	Another index considered by \cite{konapalaExploringSentinel1Sentinel22021} is the \emph{\acrfull{awei}} (\cite{feyisaAutomatedWaterExtraction2014, konapalaExploringSentinel1Sentinel22021}). This index is based upon an empirical analysis of satellite imagery in order to facilitate good separation of the water and non-water classes. In order to handle cloud shadows, \cite{feyisaAutomatedWaterExtraction2014} also provide the \acrfull{aweish}.
	\begin{align}
		\text{\acrshort{awei}} = & 4 \cdot (\text{GREEN} - \text{\acrshort{swir}-1}) \notag\\&- \fracinv{4} (\text{\acrshort{nir}} + 11 \cdot \text{\acrshort{swir}-2}) \\
		\text{\acrshort{aweish}} = & \text{BLUE} + \frac{5}{2} \cdot \text{GREEN} \notag\\&- \frac{3}{2} \cdot(\text{\acrshort{nir}} + \text{\acrshort{swir}-1}) - \frac{\text{\acrshort{swir}-2}}{4}
	\end{align}
	As is also done by \cite{konapalaExploringSentinel1Sentinel22021}, the concatenation along the channel axis of \acrshort{ndwi} and \acrshort{mndwi} is referred to as \emph{\acrshort{cndwi}}. Similarly, \emph{\acrshort{cawei}} refers to the concatenation of \acrshort{awei} and \acrshort{aweish}.
	
	Furthermore, the \acrfull{hsv} transformation \cite{smithColorGamutTransform1978}, of the \acrshort{swir}-2, \acrshort{nir}, and red channels are exceedingly successful at extracting surface water (\cite{pekelHighresolutionMappingGlobal2016, konapalaExploringSentinel1Sentinel22021}). \acrshort{hsv} separates the color and intensity information present in the input channels, which may be the reason for the success of this transformation as pointed out by \cite{pekelHighresolutionMappingGlobal2016}. This feature space and its raw bands are referred to as \emph{\acrshort{hsv}(O3)} and \emph{O3} respectively.
	
	Lastly some experiments are carried out on the following additional feature spaces:
	\begin{itemize}
		\item \emph{S2} consists of all Sentinel-2 channels, whereas \emph{OPT} removes the three 60m resolution bands similar to \cite{landuytFloodMappingVegetated2020}.
		\item \emph{RGB} and \emph{RGBN} consist of the raw Red, Green, Blue, and optionally \acrlong{nir} channels. Good results on these feature spaces would  enable the use of the constructed flood mapping approaches on satellite constellations utilizing micro-satellites. Here, only a subset of the channels used by larger satellites such as Sentinel-2 is provided, but with very frequent observation of the ground surface. As a result, classifiers that could work sufficiently well with this feature space would be very suitable for emergency relief operations (\cite{mateo-garciaGlobalFloodMapping2021}).
		\item \emph{\acrshort{hsv}(RGB)} builds upon the idea of Pekel et al. that the \acrshort{hsv}-color space transformation helps separate the color information necessary for flood detection from the absolute intensity values (\cite{pekelHighresolutionMappingGlobal2016}) and performs a traditional \acrshort{hsv} decomposition on the \emph{RGB} feature space.
	\end{itemize}
	Combinations of the used feature spaces are depicted by concatenating their names with a $+$ sign in between, similar to the notation used by \cite{konapalaExploringSentinel1Sentinel22021}. However for larger combinations, it can be hard to identify the use of different data sources, so that an additional $\_$ is used to separate optical and \acrshort{sar} features. For example the string ``SAR$\_$HSV(O3)+cAWEI+cNDWI'' refers to the channel axis concatenation of \acrshort{sar}-data with optical data, where the optical data is provided as the combination of the \acrshort{hsv}(O3), \acrshort{cawei} and \acrshort{cndwi} feature spaces.
	
	To summarize, for the purpose of this paper, the following 23 feature spaces were selected to compare the individual classifiers:
	\begin{itemize}
		\item \acrshort{sar} data only feature spaces: \acrshort{sar}
		\item Optical data only feature spaces: OPT, O3, S2, RGB, RGBN, HSV(RGB), HSV(O3), cNDWI, cAWEI, cAWEI+cNDWI, HSV(O3)+cAWEI+cNDWI
		\item \acrshort{sar} and optical feature spaces: SAR$\_$OPT, SAR$\_$O3, SAR$\_$S2, SAR$\_$RGB, SAR$\_$RGBN, SAR$\_$HSV(RGB), SAR$\_$HSV(O3), SAR$\_$cNDWI, SAR$\_$cAWEI, SAR$\_$cAWEI+cNDWI, SAR$\_$HSV(O3)+cAWEI+cNDWI
	\end{itemize}
	
	\subsection{Models}\label{sec:classifiers:models} 
	\begin{table*}
		\centering
		\caption[Choices for the maximum number of leaves of \acrshort{gb_dt} for the tested feature spaces]{Choices for the maximum number of leaves of \acrshort{gb_dt} for the tested feature spaces. The feature spaces are categorized by their dimensionality as higher-dimensional input features permit more-splits (and therefore more leaves) per tree.}
		\resizebox{\textwidth}{!}{
			\begin{tabular}{ccc}
				\toprule
				\textbf{Feature Spaces} & \textbf{Dimensionality} & \textbf{Choices for the maximum number of Leaves}  \\\midrule[\heavyrulewidth]
				\acrshort{sar}, \acrshort{cndwi},\acrshort{cawei} & 2 & 2, 4 \\ 
				O3, \acrshort{rgb}, \acrshort{hsv}(\acrshort{rgb}), \acrshort{hsv}(O3) & 3 & 4, 8  \\ 
				\acrshort{sar}$\_$\acrshort{cndwi}, \acrshort{sar}$\_$\acrshort{cawei}, RGBN, \acrshort{cawei}+\acrshort{cndwi} & 4 & 4, 8, 16  \\ 
				\acrshort{sar}$\_$O3, \acrshort{sar}$\_$\acrshort{rgb}, \acrshort{sar}$\_$\acrshort{hsv}(\acrshort{rgb}), \acrshort{sar}$\_$\acrshort{hsv}(O3) & 5 & 8, 16, 32  \\ 
				\acrshort{sar}$\_$\acrshort{rgb}N, \acrshort{sar}$\_$\acrshort{cawei}+\acrshort{cndwi}, \acrshort{hsv}(O3)$\_$\acrshort{cawei}+\acrshort{cndwi} &  6-7 & 16, 32, 64\\ 
				OPT, S2, \acrshort{sar}$\_$OPT, \acrshort{sar}$\_$S2, \acrshort{sar}$\_$\acrshort{hsv}(O3)+\acrshort{cawei}+\acrshort{cndwi} & $> 7$ & 32, 64, 128 \\\bottomrule
			\end{tabular}
		}
		\label{tab:gbdt_leave_map}
	\end{table*}
	On these feature spaces, five classical machine learning models are evaluated. This includes a linear model trained with stochastic gradient descent, three Bayesian classifiers with different model assumptions and a gradient boosting model.
	
	To enable a fair comparison of the individual classifiers, a grid-search based hyperparameter optimization was performed for each of these feature spaces. The following Section \ref{sec:classifiers:models:simple} therefore briefly explains the linear \acrshort{sgd} model as well as the Bayesian classifiers and the hyperparameters considered in these models, before Section \ref{sec:classifiers:models:gbdt} explains the \acrshort{gb_dt} model in detail.
	\subsubsection{Linear and Bayesian Models} \label{sec:classifiers:models:simple}
	The linear model is sklearn's SGDClassifier (\cite{pedregosaScikitlearnMachineLearning2011}). This implements a linear model trained with non-batchwise stochastic gradient descent using a reduce on plateau learning rate schedule.  Hyperparameters considered were loss, in the form of huber loss, logistic loss, or hinge loss (\cite{hastieSupportVectorMachines2009}), the strength of $L_2$ regularisation $\alpha \in \big\{10^{-i} | i \in \{0, 1, 2, 3, 4\}\big\}$ and whether class re-balancing should be applied.
	
	The considered bayesian classifiers are Gaussian Naive Bayes, \acrfull{lda} and \acrfull{qda}, also in the form of implementations by the sklearn machine-learning package  (\cite{pedregosaScikitlearnMachineLearning2011}). All three models determine the posterior probability of a given class and feature vector by using Bayes Rule to compute it from the likelihood of the feature vector for that class and the computed class prior probability. The models differ here in their modeling assumptions for the likelihood of the features. Whilst Naive Bayes assumes that all features are independent of each other and accordingly uses univariate Gaussian distributions, \acrshort{lda} and \acrshort{qda} compute a multivariate Gaussian distribution for each class, with \acrshort{lda} using the same covariance matrix for all classes (\cite{hastieLinearMethodsClassification2009}). 
	
	The hyperparameters considered were the shrinkage $\rho \in \left\{ \frac{i}{10}\bvert i\in\{0, \dots, 10 \}\right\}$ for \acrshort{lda} and sklearns covariance regularisation parameter in the range $\{0.0, 0.00001, 0.0001, 0.001, 0.01, 0.1, 0.5, 1, 2, 4, 8, 10\}$ for \acrshort{qda}. Note that naive bayes does not have any hyperparameters to tune.
	
	We observe that for all models, performance is very stable across different choices of regularisation parameters. Interestingly, we observed that the best performance for the simple linear model decreased when applying class-rebalancing, despite the model being trained on a highly imbalanced dataset. For all models it was observed that the selection of the feature space is the most important parameter, as will be analyzed in detail for the best model (\acrshort{gb_dt}) in the following sections. 
	
	\subsubsection{\acrfull{gb_dt}} \label{sec:classifiers:models:gbdt}
	\begin{figure*}
		\centering
		\includegraphics[width=0.69\linewidth]{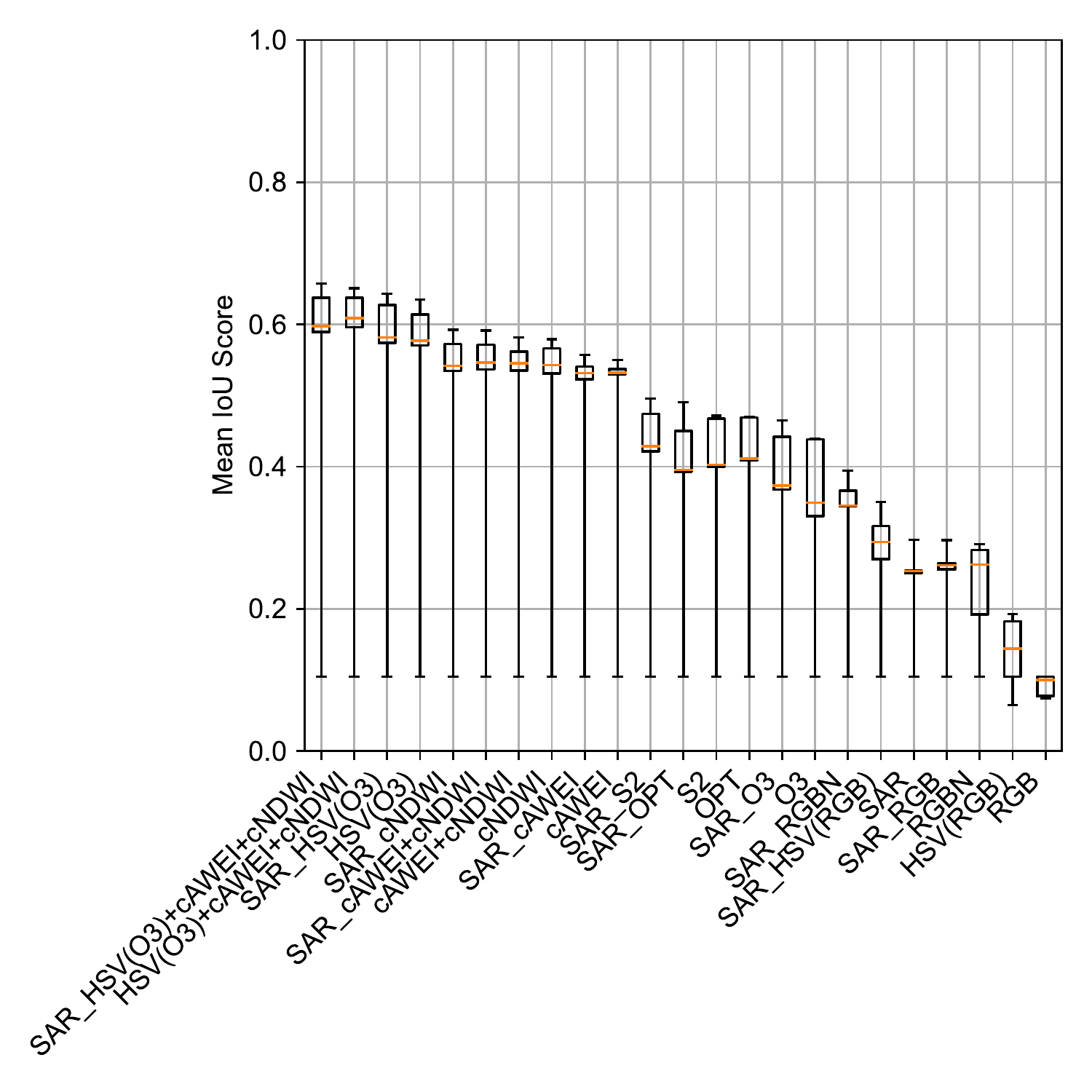}
		\caption{Influence of the used feature space on \acrshort{gb_dt} in terms of \acrsmean{iou} on the Sen1Floods11 validation set. Each box represents the results for all evaluated hyperparameters of the corresponding feature space on the x-Axis, with results for the same hyperparmeter configuration being averaged over four seeds. The x-Axis is ordered by the respective max-values, with whiskers depicting the corresponding min-max ranges.}
		\label{fig:classifiers:models:feature_spaces}
	\end{figure*}
	\begin{table*}
		\centering
		\caption[Best results reported on the Sen1Floods11 dataset]{Best results on the Sen1Floods11 test splits. The selected methods use both available data modalities, therefore Sentinel-1 (\acrshort{sar}) data and Sentinel-2 (optical) data. An extended version of this table, together with results on each modality separately, is provided in \ref{sec:appendix:quantitative}. Metric values from related work on the Sen1Floods11 dataset are taken from the individual papers and, if available, ranked by \acrstotal{iou} and otherwise \acrsmean{iou} within the evaluation split to compare them with the proposed classical machine learning approaches. The independently identically distributed (IID) split hereby refers to the regular Sen1Floods11 test set, whereas the domain shifted split refers to Bolivia test split. For each method, the used feature space as well as other parameter choices (such as hyperparameters or whether additional data has been used) are provided for a more in-depth comparison. Note that the reported standard deviation corresponds to the variation across the test set images, and not to the variation across individual runs with different seeds.}
		\resizebox{\textwidth}{!}{
			\begin{tabular}{cccccccccccc}
				\toprule %
				\multirowcell{2}{Test Split} & \multirowcell{2}{Method} & \multirowcell{2}{Feature Space} & \multirowcell{2}{Mean \acrshort{iou}\\flooded (std)} &  \multirowcell{2}{Total \acrshort{iou}\\flooded} & \multirowcell{2}{Mean \acrlong{acc}\\(std)} & \multirowcell{2}{Total \acrlong{acc}} & \multirowcell{2}{Total \acrlong{pr}\\flooded} & \multirowcell{2}{Total \acrlong{re}\\flooded} & \multirowcell{2}{Total \acrlong{re}\\dry} & \multirowcell{2}{Total \acrlong{f1}\\flooded} & \multirowcell{2}{Parameter choices}\\ 
				& & & & & & & & & & & \\ \midrule
				
				\multirowcell{13}{\emph{IID}\\\emph{Split}} & \multirowcell{5}{\acrshort{gb_dt}\\(ours)} & \multirowcell{5}{SAR$\_$HSV(O3)+cAWEI+cNDWI} & \multirowcell{5}{$\mathbf{0.7031} (\pm 0.2984)$} & \multirowcell{5}{$\mathbf{0.8751}$} & \multirowcell{5}{$\mathbf{0.9718} (\pm 0.1176)$} & \multirowcell{5}{$\mathbf{0.9838}$} & \multirowcell{5}{$\mathbf{0.9577}$} & \multirowcell{5}{$0.9103$} & \multirowcell{5}{$\mathbf{0.9943}$} & \multirowcell{5}{$\mathbf{0.9334}$} & $200$ trees\\ 
				& & & & & & & & & & & up to $128$ leaves per tree\\
				& & & & & & & & & & & regularisation $\lambda=1$ \\ 
				& & & & & & & & & & & learning rate: $0.1$ \\ 
				& & & & & & & & & & & subsample size: $262144$ \\ \cmidrule{2-12}
				
				& \multirowcell{3}{Naive\\Bayes\\(ours)} & \multirowcell{3}{SAR$\_$HSV(O3)+cAWEI+cNDWI} & \multirowcell{3}{$0.5187 (\pm 0.3168)$} & \multirowcell{3}{$0.7399$} & \multirowcell{3}{$0.9522 (\pm 0.0860)$} & \multirowcell{3}{$0.9586$} & \multirowcell{3}{$0.7784$} & \multirowcell{3}{$\mathbf{0.9378}$} & \multirowcell{3}{$0.9616$} & \multirowcell{3}{$0.8503$} & \multirowcell{3}{-}\\ 
				& & & & & & & & & & & \\ 
				& & & & & & & & & & & \\ \cmidrule{2-12}
				
				& \multirowcell{2}{\cite{yadavAttentiveDualStream2022}} & \multirowcell{2}{SAR} & \multirowcell{2}{-} & \multirowcell{2}{$0.70$} & \multirowcell{2}{-} & \multirowcell{2}{-} & \multirowcell{2}{-} & \multirowcell{2}{-} & \multirowcell{2}{-} & \multirowcell{2}{0.83} & Attentive Dual Stream Siamese Network \\ 
				& & & & & & & & & & & Additional Pre-Flood \acrshort{sar} Data\\  \cmidrule{2-12}
				
				& \multirowcell{3}{\cite{baiEnhancementDetectingPermanent2021}} & \multirowcell{3}{SAR$\_$S2} & \multirowcell{3}{$0.5873$} & \multirowcell{3}{$0.6452$} & \multirowcell{3}{$0.9338$} & \multirowcell{3}{-} & \multirowcell{3}{-} & \multirowcell{3}{$0.6881$} & \multirowcell{3}{$0.9884$} & \multirowcell{3}{-} & BASNet (\cite{qinBASNetBoundaryAwareSalient2019}) \\ 
				& & & & & & & & & & & \multirowcell{2}{Uses Sen1Floods11 weakly-labeled\\split for pretraining (4384 images)} \\ 
				& & & & & & & & & & &\\ \midrule
				
				\multirowcell{11}{\emph{Domain}\\\emph{Shifted}\\\emph{Split}} & \multirowcell{5}{\acrshort{gb_dt}\\(ours)} & \multirowcell{5}{SAR$\_$HSV(O3)+cAWEI+cNDWI} & \multirowcell{5}{$\mathbf{0.6070} (\pm 0.3542)$} & \multirowcell{5}{$\mathbf{0.8357}$} & \multirowcell{5}{$\mathbf{0.9705} (\pm 0.0293)$} & \multirowcell{5}{$\mathbf{0.9730}$} & \multirowcell{5}{$\mathbf{0.9584}$} & \multirowcell{5}{$0.8671$} & \multirowcell{5}{$\mathbf{0.9929}$} & \multirowcell{5}{$\mathbf{0.9105}$} & $200$ trees \\ 
				& & & & & & & & & & & up to $128$ leaves per tree\\ 
				& & & & & & & & & & & regularisation $\lambda=1$\\ 
				& & & & & & & & & & & learning rate: $0.1$\\ 
				& & & & & & & & & & & subsample size: $262144$\\ \cmidrule{2-12}
				
				& \multirowcell{3}{\cite{baiEnhancementDetectingPermanent2021}} & \multirowcell{3}{SAR$\_$S2} & \multirowcell{3}{$0.5407$} & \multirowcell{3}{$0.7890$} & \multirowcell{3}{$0.9579$} & \multirowcell{3}{-} & \multirowcell{3}{-} & \multirowcell{3}{$0.9234$} & \multirowcell{3}{$0.9679$} & \multirowcell{3}{-} & BASNet (\cite{qinBASNetBoundaryAwareSalient2019}) \\ 
				& & & & & & & & & & & \multirowcell{2}{Uses Sen1Floods11 weakly-labeled\\split for pretraining (4384 images)} \\ 
				& & & & & & & & & & &\\ \cmidrule{2-12}
				
				& \multirowcell{3}{Naive\\Bayes\\(ours)} & \multirowcell{2}{SAR$\_$HSV(O3)+cAWEI+cNDWI} & \multirowcell{3}{$0.4402 (\pm 0.2889)$} & \multirowcell{3}{$0.6160$} & \multirowcell{3}{$0.8984 (\pm 0.0776)$} & \multirowcell{3}{$0.9033$} & \multirowcell{3}{$0.6268$} & \multirowcell{3}{$\mathbf{0.9728}$} & \multirowcell{3}{$0.8902$} & \multirowcell{3}{$0.7620$} & \multirowcell{2}{-}\\ 
				& & & & & & & & & & & \\ 
				& & & & & & & & & & & \\ 
				\bottomrule
			\end{tabular}
		}
		\label{tab:classifier_comparison_combined}
	\end{table*} 
	Random forest or gradient boosting based models have been applied effectively for flood inundation mapping before (\cite{leeSpatialPredictionFlood2017,fengFloodMappingBased2015,delacruzNearRealtimeFloodDetection2020}). Gradient boosting is a steepest-descent ensemble learning method, fitting a model $f(x)$ of $M$ weak base learners $b$ (characterized by input $\mvec{x}$ and parameters $\gamma_m, 1\leq m \leq M$) to form an additive function with expansion coefficients $\beta_m, 1\leq m \leq M$ (\cite{hastieBoostingAdditiveTrees2009}):
	\begin{equation}\label{eq:gradient_boosting:model}
		f(\mvec{x}) = \sum\limits_{m=1}^M\beta_m b(\mvec{x},\gamma_m)
	\end{equation}
	In our case, only decision trees are considered as weak learners $b$ due to their good performance on other remote sensing tasks, to form the \acrlong{gb_dt} model. It is left to future work, to consider other valid weak learners, such as support vector machines (\cite{hastieBoostingAdditiveTrees2009,friedmanGreedyFunctionApproximation2001}).
	
	To fit the model in equation \ref{eq:gradient_boosting:model}, the optimization problem in equation \ref{eq:gradient_boosting:objective} has to be solved for a given loss function $\ell$, which is rarely feasible in practice:
	\begin{equation} \label{eq:gradient_boosting:objective}
		\argmin_{\substack{\beta_1, \dots, \beta_M\\\gamma_1, \dots, \gamma_M}} \sum_{i=1}^N \ell\left(y_i, \sum_{m=1}^M \beta_m b(\mvec{x_i}, \gamma_m)\right)
	\end{equation}
	Instead ``Forward Stagewise Additive Modeling'' is performed, which approximates this solution by fitting the base learners sequentially, leaving the parameters of earlier learners unmodified (\cite{hastieBoostingAdditiveTrees2009}). It can be shown, that the optimal choice of parameters for model $m$ corresponds to the closest fit to the negative gradient of model $m-1$ to the loss $\ell$ (\cite{friedmanGreedyFunctionApproximation2001}), resulting in the name ``gradient boosting''.
	
	As gradient boosting implementation, the highly-scalable implementation provided by LightGBM (\cite{keLightGBMHighlyEfficient2017}) is used, which fits leaf-wise grown decision trees as weak base learners. Hereby all current leaves are considered as expansion candidates in each tree-building step, whereas traditional level-wise expansions first expand all nodes in each level, before considering leaves from the next level (\cite{shiBestfirstDecisionTree2007, lightgbm-teamFeaturesLightGBM99}). This allows for fewer nodes per-tree to achieve a given accuracy, at the cost of potentially causing over-fitting if the number of data points is small, however as there are more than $50M$ pixels in the hand-labeled training-set this is highly unlikely. 
	
	As is also pointed out by the LightGBM documentation, the number of boosting iterations performed and the number of leaves per tree are therefore very important parameters to tune. Higher values for both, allow the ensemble to better fit the training set, but may cause overfitting (\cite{lightgbm-teamFeaturesLightGBM99, lauraeLauraeXgboostLightGBM}). As expected, increasing the number of boosting iterations performed (and therefore the number of trees in the model), does improve performance, however this effect is rather insignificant and there are diminishing returns.
	
	LightGBM also offers class re-balancing for training the forests, but similar to the linear \acrshort{sgd} model, this does not improve the performance. Instead we can only observe a smaller \acrfull{iqr} and value range, indicating that the model is more stable with respect to the tested hyper-parameters, but does not converge towards the best solution achievable without re-balancing. We provide additional figures, depicting both effects, in \ref{sec:appendix:hyperparameters}.
	
	Due to the individual feature spaces varying in dimensionality ($2 \leq d \leq 15$), deeper trees (trees with more nodes) are better suited to some feature spaces, than to others. Therefore the number of maximum leaves to test was hand-selected depending on the dimensionality as given in Table \ref{tab:gbdt_leave_map}. Therefore, it is of higher interest to inspect the corresponding feature spaces, than the number of leaves used. 
	
	\section{Results} \label{sec:results}
	\textbf{The Feature Space is the single most important parameter to tune.} This is shown in Figure \ref{fig:classifiers:models:feature_spaces}, which depicts the performance of the classifier in terms of \acrsmean{iou} with respect to the choice of the feature space. We find that the best feature space (SAR$\_$HSV(O3)+cAWEI+cNDWI), significantly outperforms the worst feature space (\acrshort{rgb}) with a validation set \acrsmean{iou} of $0.6574$ over $0.1047$. Similar differences could be observed for the other considered machine learning classifiers on their respective best and worst performing feature spaces, which can be found in \ref{sec:appendix:hyperparameters}.
	
	Furthermore, the hand-crafted feature spaces consistently outperform band combinations made up only of raw Sentinel-1 or Sentinel-2 data (S2, \acrshort{rgb}, \acrshort{sar}, ...). All feature spaces containing transformations appear before the raw band combinations in the ranking of Figure \ref{fig:classifiers:models:feature_spaces}. Additionally, we find that the \acrshort{gb_dt} model is able to utilize multiple modalities as long as the features are of sufficient quality. The models using both \acrshort{sar} and optical data always outperform those trained on optical or SAR data only. Accordingly, \acrshort{gb_dt} models can also provide successful baselines if additional modalities become available on the Sen1Floods11 dataset.
	
	However, the performance of the considered machine learning models using only \acrshort{sar} or \acrshort{rgb}-data is far from enabling a satisfactory detection of flooded surface water. Although the \acrshort{hsv}-transformation of the red, green and blue bands proposed in Section \ref{sec:data:feature_spaces} is helpful, the observed performance is still unsatisfactory. To achieve a successful delineation of flooded areas, the multispectral channels of Sentinel-2 are essential.
	\subsection{Quantitative Analysis}
	\label{sec:classifiers:quantitative}
	\begin{table*}
		\caption[Regionwise Total Evaluation]{Statistics of per-region total metrics, calculated for the \acrshort{gb_dt} classifier on the Sen1Floods11 test set. In order to provide an estimate of the \acrshort{gb_dt}-model's performance on regions all around the world, mean, standard deviation, value range and median of the per-region evaluations on the Sen1Floods11 test set are reported.}
		\resizebox{\textwidth}{!}{
			\begin{tabular}{ccccccc}\toprule
				Feature Space & Statistic & \acrshort{acc} & \acrshort{iou} & \acrlong{f1} & \acrlong{pr} & \acrlong{re} \\ \midrule
				\multirowcell{4}{\acrshort{sar}$\_$HSV(O3)+cAWEI+cNDWI} & Mean (std.) & $\mathbf{0.9857} (\pm 0.0125)$ & $\mathbf{0.7998} (\pm 0.1750)$ & $\mathbf{0.8788} (\pm 0.1141)$ & $0.9372 (\pm 0.0626)$ & $\mathbf{0.8404} (\pm 0.1673)$ \\ \cmidrule{2-7} 
				& Min-Max & $\mathbf{0.9612 - 0.9972}$ & $\mathbf{0.5468 - 0.9713}$ & $\mathbf{0.7070 - 0.9855}$ & $0.7829 - 0.9880$ & $\mathbf{0.5618 - 0.9832}$ \\ \cmidrule{2-7}
				& Median & $\mathbf{0.9912}$ & $\mathbf{0.8888}$ & $\mathbf{0.9410}$ & $0.9561$ & $\mathbf{0.9189}$ \\ \midrule
				\multirowcell{4}{HSV(O3)+cAWEI+cNDWI} & Mean (std.) & $0.9856 (\pm 0.0125)$ & $0.7925 (\pm 0.1876)$ & $0.8723 (\pm 0.1261)$ & $\mathbf{0.9406 (\pm 0.0581)}$ & $0.8311 (0.1844)$ \\ \cmidrule{2-7} 
				& Min-Max & $0.9616 - 0.9968$ & $0.4684 - 0.9754$ & $0.6380 - 0.9876$ & $\mathbf{0.8035 - 0.9920}$ & $0.4749 - 0.9832$ \\ \cmidrule{2-7}
				& Median & $0.9910$ & $0.8860$ & $0.9394$ & $\mathbf{0.9647}$ & $0.9162$ \\ \midrule
				\multirowcell{4}{\acrshort{sar}} & Mean (std.) & $0.9317 (\pm 0.0432)$ & $0.4054 (\pm 0.1960)$ & $0.5517 (\pm 0.2019)$ & $0.7569 (\pm 0.2529)$ & $0.4883 (\pm 0.2013)$ \\ \cmidrule{2-7} 
				& Min-Max & $0.8448 - 0.9928$ & $0.1457 - 0.6819$ & $0.2544 - 0.8109$ & $0.2527 - 0.9775$ & $0.1531 - 0.6928$ \\ \cmidrule{2-7}
				& Median & $0.9409$ & $0.3912$ & $0.5622$ & $0.8215$ & $0.5375$ \\ 
				\bottomrule
			\end{tabular}
		}
		\label{tab:region_wise_performance}
	\end{table*}

	To allow for a quantitative comparison with related work, Table \ref{tab:classifier_comparison_combined} shows the test and bolivia test set, described in Section \ref{sec:classifiers:metrics}, results of the evaluated \acrshort{gb_dt} model and the naive bayes classifier combined \acrshort{sar} and optical data. Additional results of all tested models and evaluations using both data modalities individually are provided in \ref{sec:appendix:quantitative}. The depicted methodologies were selected on the validation set primarily according to their mean \acrshort{iou} score, with the other metrics functioning as tie breakers in the order they appear in the table. 
	\newpage
	\textbf{All considered classifiers outperform previous state-of-the-art neural network based approaches} on the Sen1Floods11 test set in terms of \acrstotal{iou}, if multispectral data is available. We find that the selected \acrshort{gb_dt} model consistently outperforms all other methods for all metrics, with the exception of \acrltotal{re}, for which the Naive Bayes classifier consistently performs best. \acrshort{gb_dt} achieves a test set mean and \acrstotal{iou} of $0.7010$ and $0.8767$ using optical data alone and $0.7031$, $0.8751$ using both optical and \acrshort{sar} data. 
	
	As mentioned previously in Section \ref{sec:related:ml}, \cite{yadavAttentiveDualStream2022} provide the current best \acrstotal{iou} of $0.70$ on the Sen1Floods11 dataset. Whilst they do not provide \acrsmean{iou} values to compare to, it is observed in Table \ref{tab:classifier_comparison_combined}, that the \acrshort{gb_dt} model exceeds this with an improvement of $0.1751$ and $25.01\%$ in absolute and relative terms. 
	
	However, it has to be noted that both models use different datasources. Whereas our \acrshort{gb_dt} model uses optical and \acrshort{sar} post-flood data, \cite{yadavAttentiveDualStream2022} use pre-flood and post-flood \acrshort{sar}-Data to perform change detection, so that a comparison of both methods is hard. The best-performing approach using the same data sources as our model, though with more training data, is the approach by \cite{baiEnhancementDetectingPermanent2021}. Our \acrshort{gb_dt} model exceeds this methodology in terms of mean and total \acrshort{iou} with a relative improvement of $19.7\%$ and $35.63\%$.
	
	The previously reported neural network based models only exceed \acrshort{gb_dt} for the flooded \acrlong{re} on the bolivia test set. This is due to the BASNet used by \cite{baiEnhancementDetectingPermanent2021} achieving an \acrshort{oe} of $0.0766$ which corresponds to a flooded \acrlong{re} of $0.9234$, which is larger than the $0.8671$ achieved by \acrshort{gb_dt}. On the other hand, in terms of flooded \acrlong{re} the naive bayes classifier consistently outperforms all other methods, including the approach by \cite{baiEnhancementDetectingPermanent2021}, thereby showing that for a suitable choice of feature space and classical machine learning method, the state-of-the-art neural network based approaches can be exceeded.
	
	In general, the tested models tend to perform worse on the Bolivia test set, than on the regular test set. The only exception to this, is the linear model trained with \acrshort{sgd} for \acrshort{sar} data that achieves a mean \acrshort{iou} of $0.4202$ on the Bolivia test set whilst achieving a mean \acrshort{iou} of $0.2597$ on the regular test set. This is a particularly surprising result, as all considered methodologies do not achieve a satisfying performance using \acrshort{sar}-only data on the test set. Here the best observed result is again the \acrshort{gb_dt} model with a mean \acrshort{iou} of $0.2880$ with the second best being the linear model. 
	\newpage
	\textbf{The models do not perform very consistent across the individual regions}. This is already indicated by the difference between the test and bolivia test sets, however to further analyze this effect, the standard deviation for mean-based metrics across the evaluation datasets was inspected as well and is presented in Table \ref{tab:classifier_comparison_combined}. We find that the \acrshort{gb_dt}-model exhibits a very large standard deviation so that it is likely that there are some images with very low \acrshort{iou} values. Furthermore, we find that for the other models the $2\sigma$ interval of the gaussian distribution given by the sample mean and standard deviation across the \acrshort{iou}s calculated for the individual images includes an \acrshort{iou} of $0$, as is shown in \ref{sec:appendix:quantitative}. 
	
	\begin{figure}
		\centering
		\begin{subfigure}{0.9\linewidth}
			\centering
			\includegraphics[width=0.8\linewidth]{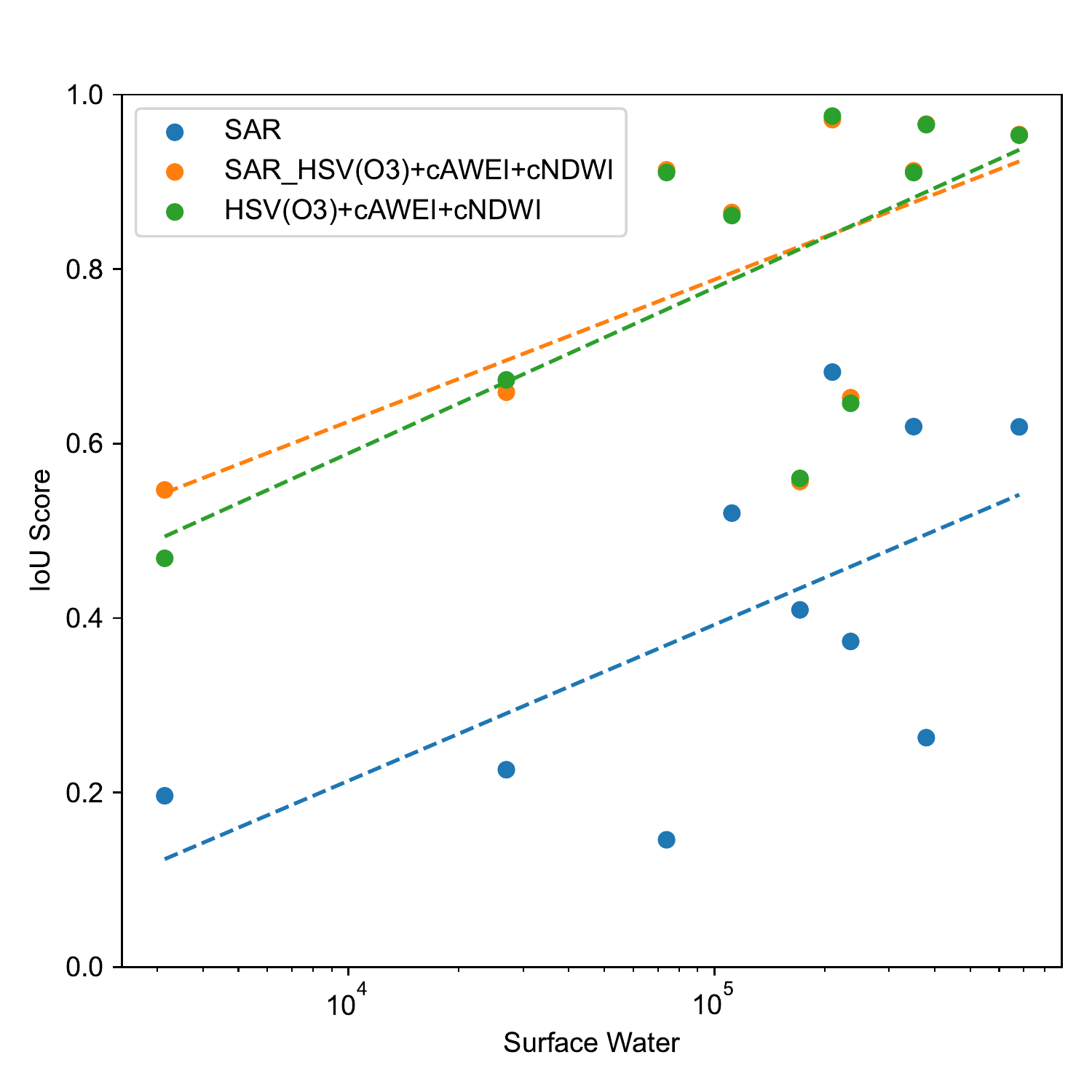}
			\caption{Total \acrshort{iou} for the surface water class}
			\label{fig:classifiers:models:region_performance:flood_iou}
		\end{subfigure}\\
		\begin{subfigure}{0.9\linewidth}
			\centering
			\includegraphics[width=0.8\linewidth]{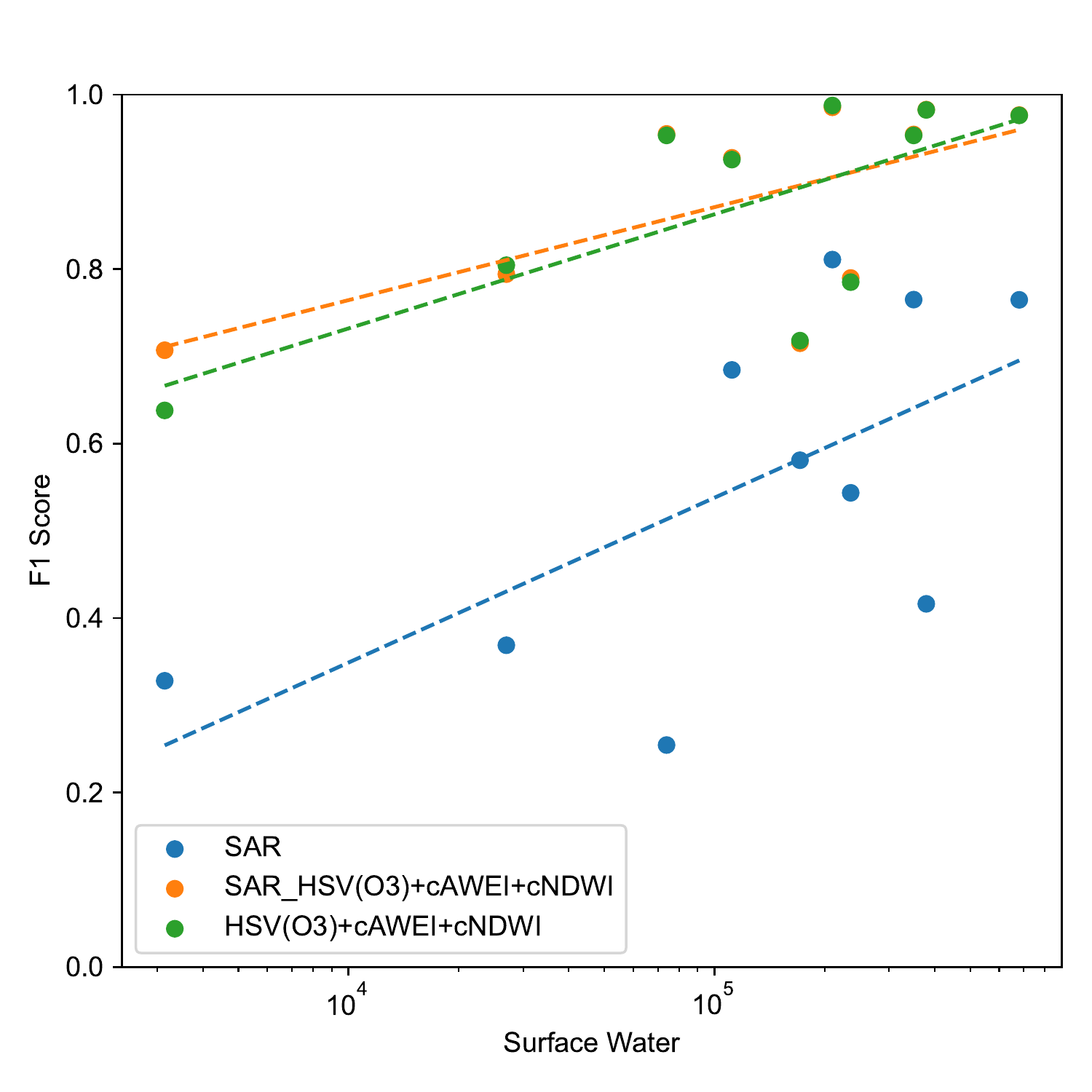}
			\caption{Total \acrlong{f1} for the surface water class}
			\label{fig:classifiers:models:region_performance:flood_f1}
		\end{subfigure}\\
		\begin{subfigure}{0.9\linewidth}
			\centering
			\includegraphics[width=0.8\linewidth]{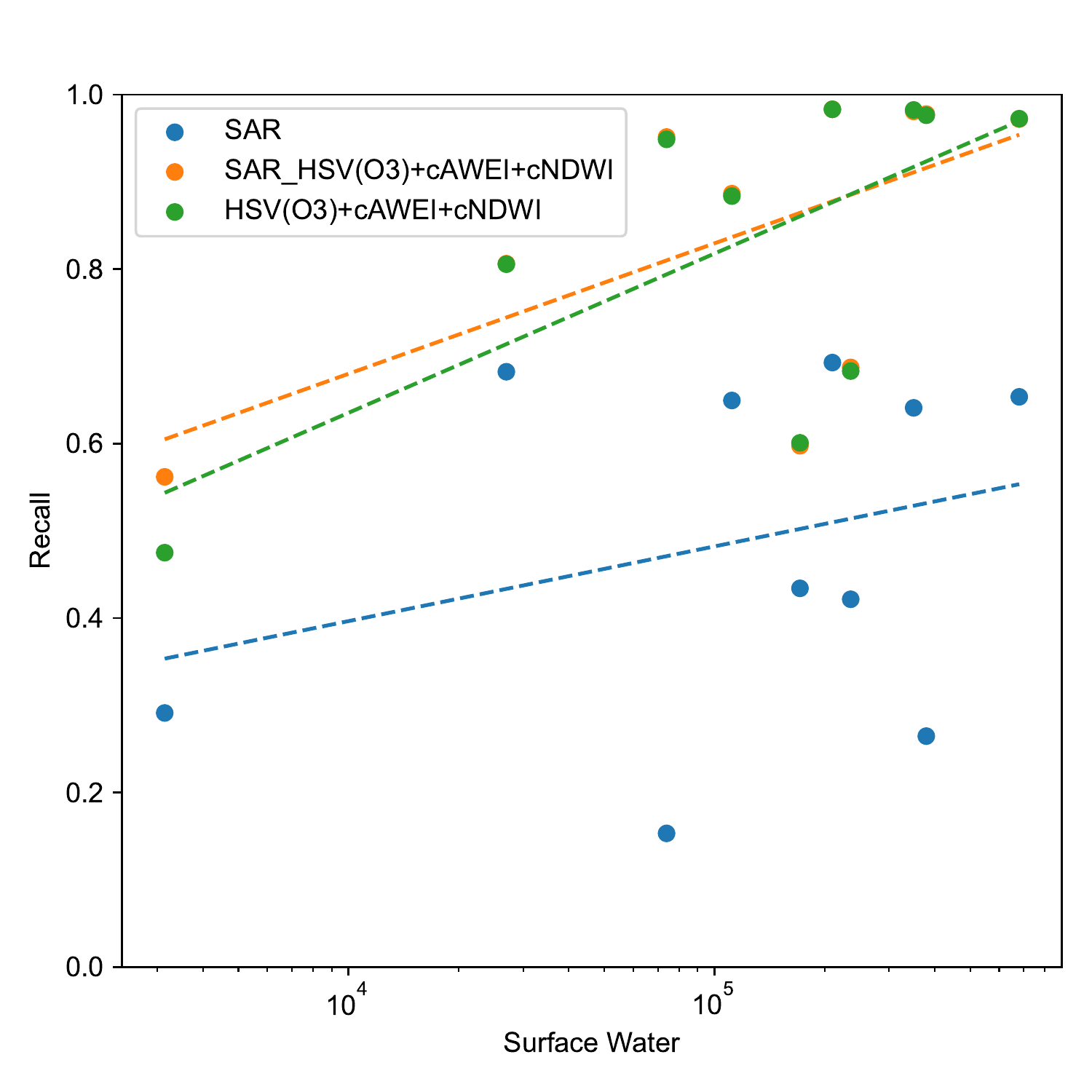}
			\caption{Total \acrlong{re} for the surface water class}
			\label{fig:classifiers:models:region_performance:flood_re}
		\end{subfigure}\\
		\caption[Performance of the \acrshort{gb_dt}-model per region.]{Total region-wise metrics for the \acrshort{gb_dt}-model plotted as scatter plots. The amount of flooded pixels in the regions is depicted on the log-scale x-axis and the corresponding metric is depicted on the y-axis. Additionally, for each model a regression line is plotted to depict the trend present in the data.}
		\label{fig:classifiers:models:region_performance}
	\end{figure}
	
	Due to region imbalance, it is difficult to achieve a good estimate of the actual performance of the classifier on arbitrary regions of the world. For instance, mean-based metrics are biased with respect to majority regions, while total metrics such as \acrstotal{iou} favor regions with a larger representation of the minority class. \cite{konapalaExploringSentinel1Sentinel22021} propose a regionwise cross-validation approach as an alternative evaluation methodology, where the model is trained ten times on nine of the eleven regions in the Sen1Floods11 dataset and then evaluated on the other two regions using total metrics. However, this does not solve the region imbalance problem as the two tested regions may be of varying size and furthermore does not allow estimating the performance of the classifier on in-distribution samples, thus requiring a different evaluation metric.
	
	To further analyze the performance of the model on a global scale, we therefore propose to use total metrics that are calculated per-region for the selected \acrshort{gb_dt} classifiers. Plotting the so created metrics on a log-scale axis for the number of water pixels within each region, we can observe that for \acrshort{iou}, \acrlong{f1} and \acrlong{re} a clear trend is present for the model on all three data modalities. In regions with a larger number of water pixels, the models perform better for the flooded class as shown in Figures \ref{fig:classifiers:models:region_performance:flood_iou}, \ref{fig:classifiers:models:region_performance:flood_f1}, and \ref{fig:classifiers:models:region_performance:flood_re}. Whilst for \acrlong{pr} of the \acrshort{sar} \acrshort{gb_dt}-model the trend is also visible, we cannot observe it there for the other data modalities. 
	
	\textbf{There exists a correlation between the amount of water pixels in a region and the classifiers performance}. To verify this statement, the python sci-analysis package (\cite{morrowchrisScianalysis2019}) was used for performing statistical significance tests using the Spearman or Pearson Correlation Coefficients as appropriate. We find statistically significant (p-value $<0.05$) correlations for \acrshort{iou} and \acrlong{f1} of the optical data only model and for \acrlong{f1} and \acrlong{pr} using the \acrshort{sar} data only model. Furthermore, only \acrlong{acc} and \acrlong{pr} result in p-values greater than $0.09$ for the optical data based models, indicating that even for the remaining metrics, a bias towards majority regions is present.
	
	\textbf{\acrshort{gb_dt} performs well all around the world, despite the regional bias}. We propose to use the mean and standard deviation of the total regionwise metrics, to allow a final quantitative evaluation of the classifier in a global context, independent of the region imbalance of the Sen1Floods11 dataset. The so calculated \acrsmean{iou} value of $0.7998$ hereby serves as an estimate of the performance of the classifier, whilst the standard deviation of $0.1750$ permits an analysis of the performance on regions all around the world. Due to the aforementioned statistically significant bias present in the model, we additionally provide the value range and median in Table \ref{tab:region_wise_performance}. With \acrshort{iou} values ranging from $0.5468$ to $0.9713$, we conclude that \acrshort{gb_dt} can provide high-quality flood inundation maps all around the world.
	
	\subsection{Qualitative Analysis} \label{sec:classifiers:qualitative}
	\begin{figure*}
		\centering
		\begin{subfigure}{0.45\linewidth}
			\centering
			\includegraphics[width=0.8\linewidth]{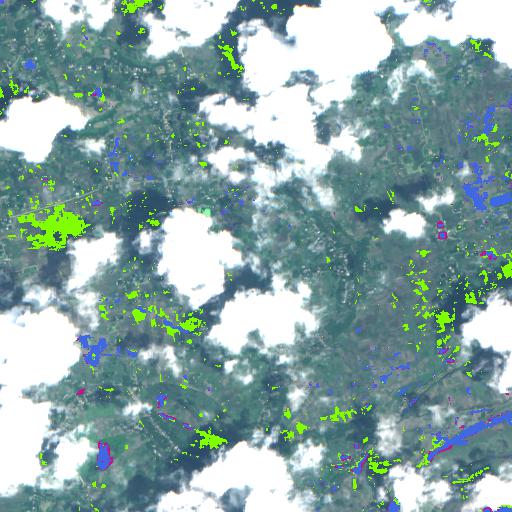}
			\caption{Image from India}
			\label{fig:classifiers:qualitative:gb_dt_negative_samples:clouds}
		\end{subfigure}
		\begin{subfigure}{0.45\linewidth}
			\centering
			\includegraphics[width=0.8\linewidth]{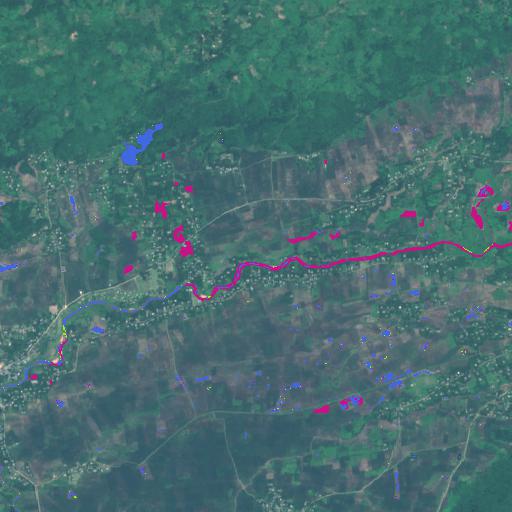}
			\caption{Image from India}
			\label{fig:classifiers:qualitative:gb_dt_negative_samples:river1}
		\end{subfigure}\\
		\begin{subfigure}{0.45\linewidth}
			\centering
			\includegraphics[width=0.8\linewidth]{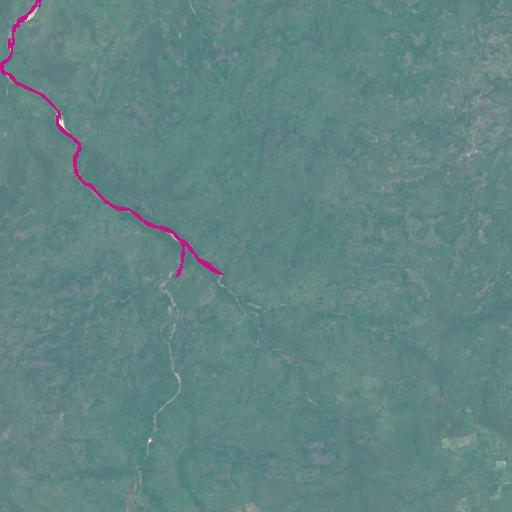}
			\caption{Image from Ghana}
			\label{fig:classifiers:qualitative:gb_dt_negative_samples:large_miss}
		\end{subfigure}
		\begin{subfigure}{0.45\linewidth}
			\centering
			\includegraphics[width=0.8\linewidth]{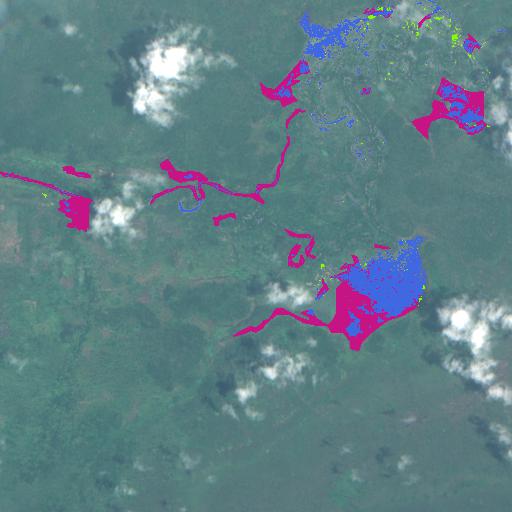}
			\caption{Image from Ghana}
			\label{fig:classifiers:qualitative:gb_dt_negative_samples:river2}
		\end{subfigure}\\
		\begin{subfigure}{0.45\linewidth}
			\centering
			\includegraphics[width=0.8\linewidth]{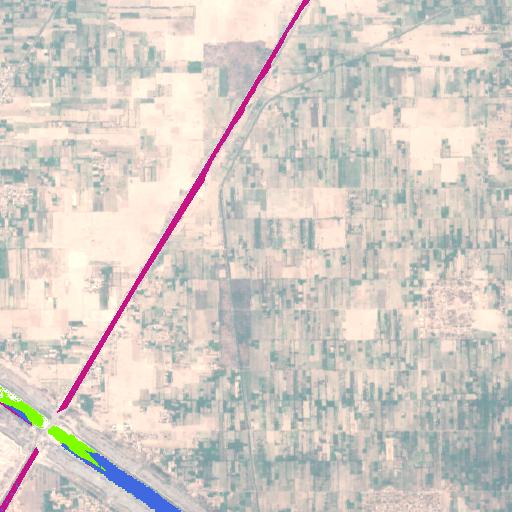}
			\caption{Image from Paraguay}
			\label{fig:classifiers:qualitative:gb_dt_negative_samples:river3}
		\end{subfigure}
		\begin{subfigure}{0.45\linewidth}
			\centering
			\includegraphics[width=0.8\linewidth]{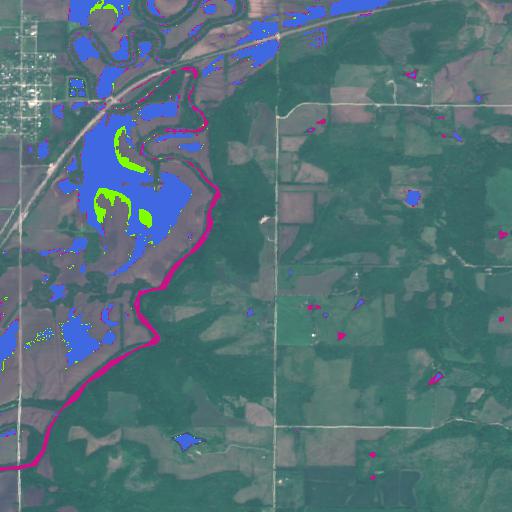}
			\caption{Image from the USA}
			\label{fig:classifiers:qualitative:gb_dt_negative_samples:river4}
		\end{subfigure}\\
		\caption[Failure cases of the \acrshort{gb_dt}-model for differing meteorological and geographical conditions]{Failure cases of the \acrshort{gb_dt}-model for differing meteorological and geographical conditions. Areas marked in blue correspond to correct flood predictions, whereas magenta and green correspond to \acrlong{fn} and \acrlong{fp} respectively.}
		\label{fig:classifiers:qualitative:gb_dt_negative_samples}
	\end{figure*}
	\begin{figure*}
		\centering
		\begin{subfigure}{0.45\linewidth}
			\centering
			\includegraphics[width=0.8\linewidth]{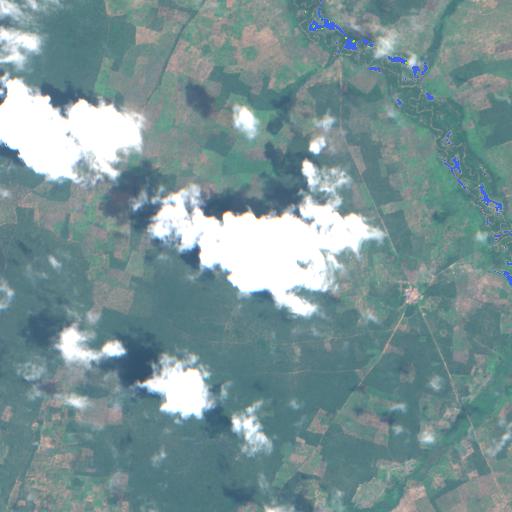}
			\caption{Image from Nigeria}
			\label{fig:classifiers:qualitative:gb_dt_positive_samples:cloudy}
		\end{subfigure}
		\begin{subfigure}{0.45\linewidth}
			\centering
			\includegraphics[width=0.8\linewidth]{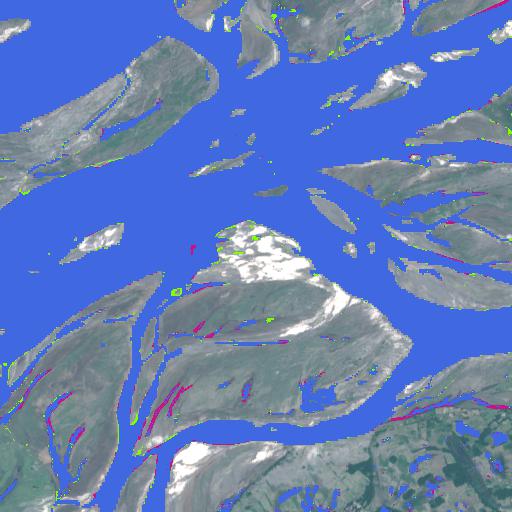}
			\caption{Image from India}
			\label{fig:classifiers:qualitative:gb_dt_positive_samples:large2}
		\end{subfigure}\\
		\begin{subfigure}{0.45\linewidth}
			\centering
			\includegraphics[width=0.8\linewidth]{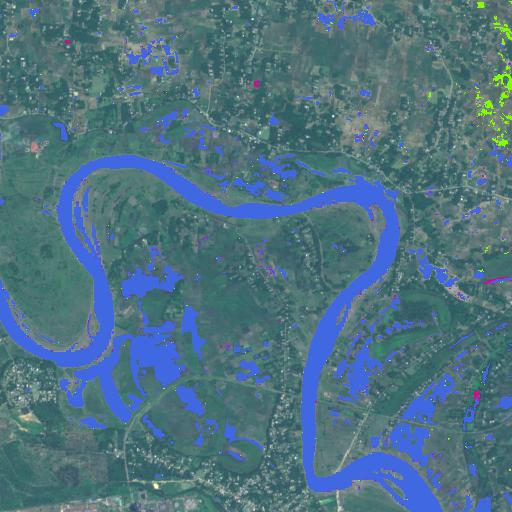}
			\caption{Image from India}
			\label{fig:classifiers:qualitative:gb_dt_positive_samples:large1}
		\end{subfigure}
		\begin{subfigure}{0.45\linewidth}
			\centering
			\includegraphics[width=0.8\linewidth]{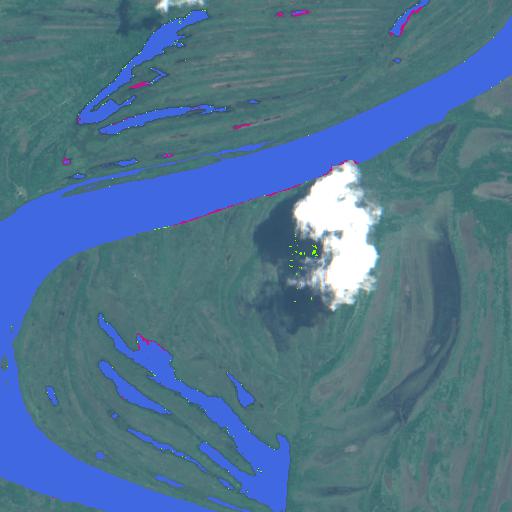}
			\caption{Image from Paraguay}
			\label{fig:classifiers:qualitative:gb_dt_positive_samples:large3}
		\end{subfigure}\\
		\begin{subfigure}{0.45\linewidth}
			\centering
			\includegraphics[width=0.8\linewidth]{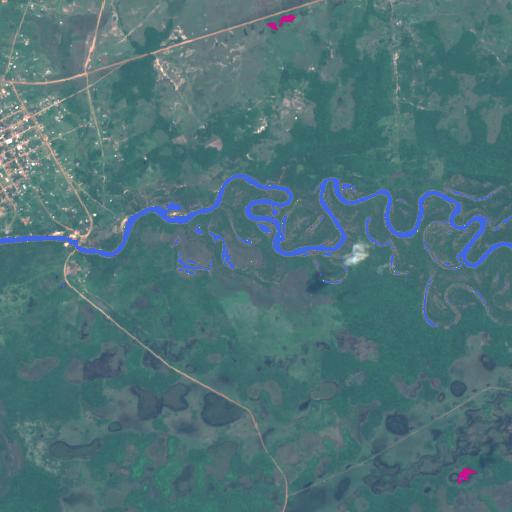}
			\caption{Image from Paraguay}
			\label{fig:classifiers:qualitative:gb_dt_positive_samples:small1}
		\end{subfigure}
		\begin{subfigure}{0.45\linewidth}
			\centering
			\includegraphics[width=0.8\linewidth]{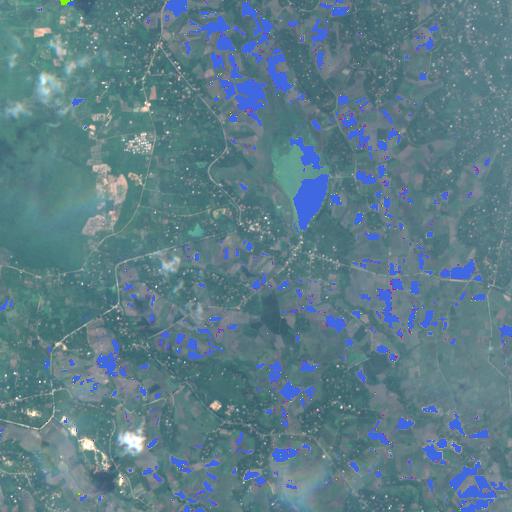}
			\caption{Image from Sri-Lanka}
			\label{fig:classifiers:qualitative:gb_dt_positive_samples:small2}
		\end{subfigure}\\
		\caption[Positive examples of the \acrshort{gb_dt}-model for differing meteorological and geographical conditions]{Positive examples of the \acrshort{gb_dt}-model for differing meteorological and geographical conditions. Areas marked in blue correspond to correct flood predictions, whereas magenta and green correspond to \acrlong{fn} and \acrlong{fp} respectively.}
		\label{fig:classifiers:qualitative:gb_dt_positive_samples}
	\end{figure*}
	\textbf{Clouds negatively impact the performance of \acrshort{gb_dt}}. Positive and negative cases of the \acrshort{gb_dt} model are inspected for a qualitative analysis of the performance of the optical data based \acrshort{gb_dt}-classifier on the Sen1Floods11 test set. As expected from a model that is mainly based on optical data, the first observation that can be made is that clouds negatively affect the resulting performance. Figure \ref{fig:classifiers:qualitative:gb_dt_negative_samples:clouds} shows that the forest-based classifier tends to erroneously predict areas affected by cloud shadows as flooded. Furthermore, in some images the classifier misses smaller riverbeds (Figures \ref{fig:classifiers:qualitative:gb_dt_negative_samples:river1}, \ref{fig:classifiers:qualitative:gb_dt_negative_samples:river2}, \ref{fig:classifiers:qualitative:gb_dt_negative_samples:river3}, \ref{fig:classifiers:qualitative:gb_dt_negative_samples:river4}) or in one case even large parts of a connected water segment (Figure \ref{fig:classifiers:qualitative:gb_dt_negative_samples:large_miss}). However, as is already shown by the high mean \acrshort{iou}, these errors are rare.
	
	Presumably, the cause for the misclassification under cloudy conditions, is that the classifier operates on a pixel-wise basis. Figure \ref{fig:classifiers:qualitative:gb_dt_positive_samples:cloudy} shows that the classifier is in principle capable of correctly recognizing dry land that is obscured by cloud shadows. However, for some objects on the ground, pixels may look identical under cloudy conditions to water pixels in other areas, so that the classifier cannot correctly identify them as dry land as it is missing the information of a cloud being close by. In this context, a windowed approach might be able to achieve better results. Alternatively, one could potentially overcome this limitation by adding the cloud shadow masks calculated for Sentinel-2 imagery by the Copernicus services and providing them as an additional input to the \acrshort{gb_dt} model.
	
	In a similar sense, the sometimes missed features are likely caused by the model being trained on the complete dataset and not in a batch-wise manner. In contrast to neural-network based approaches, that are usually trained via minibatch-\acrshort{sgd} and therefore are required to perform well on very different kinds of images to be able to reduce the training loss, the \acrshort{gb_dt}-model always observes pixels from all images in the dataset during training. Therefore, the model can learn to neglect some images in the dataset, that then might result in the mentioned negative samples. 
	
	In order to improve on these results, it may therefore be an interesting direction for future work to explore the potential of training the \acrshort{gb_dt}-model in a batched way as well, thereby leveraging the same regularisation effect. This procedure, in combination with for example a region-stratified sampling of batches, might also benefit the stability of the algorithm with respect to different regions around the world.
	\newpage	
	\textbf{\acrshort{gb_dt} provides highly precise segmentation maps}. This can be verified by inspecting the positive samples in Figure \ref{fig:classifiers:qualitative:gb_dt_positive_samples} which show the good performance also reported by the metrics. Images with large flooded areas, such as depicted in Figures \ref{fig:classifiers:qualitative:gb_dt_positive_samples:large1}, \ref{fig:classifiers:qualitative:gb_dt_positive_samples:large2}, and \ref{fig:classifiers:qualitative:gb_dt_positive_samples:large3}, can be segmented almost completely correct as well as images with only small, mostly disconnected flooded areas as shown in Figures \ref{fig:classifiers:qualitative:gb_dt_positive_samples:small1} and \ref{fig:classifiers:qualitative:gb_dt_positive_samples:small2}. 
	
	It should be noted, that especially the latter is an advantage of this pixel-based approach over reported neural-network based image segmentation methods, as neural-networks often produce blurry boundaries or miss smaller features in the resulting segmentation map (\cite{valanarasuKiUNetOvercompleteConvolutional2021, baiEnhancementDetectingPermanent2021}), which was not observed for the \acrshort{gb_dt}-model.
	
	\section{Conclusions and Future Work}
	\label{sec:conclusion}
	In this work, the potential of five classical machine learning models as pixel-wise classifiers for flood inundation mapping has been systematically investigated. Quantitative analysis based on an exhaustive hyperparameter search for 23 feature spaces each shows that all classical machine learning algorithms can outperform current state-of-the-art deep learning approaches in terms of \acrstotal{iou} with a relative improvement of up to $25.01\%$. Furthermore, a qualtitative analysis shows that the best-performing model, a pixel-wise \acrlong{gb_dt} classifier, enables highly accurate segmentation bounds with an accuracy that is outperforming all previous convolutional filter-based neural networks.
	
	Our analysis also shows that the \acrshort{gb_dt} model learns a bias with respect to regions with a greater presence of the surface water minority class. While this bias is not as pronounced as an initial examination of the standard deviation across the images of the test dataset would suggest, a trend can still be seen visually and was verified using statistical tests. Due to the observed region imbalance of the Sen1Floods11 dataset, we additionally propose to calculate regionwise total metrics to enable unbiased evaluations in a global context. The \acrshort{gb_dt} model hereby achieves a regionwise \acrsmean{iou} of $0.7998$ and whilst we cannot compare this to existing work yet, we make our code available so that future research can easily employ unbiased regionwise total evaluations as well as using our proposed feature spaces.
	
	It should be noted that in this paper, the \acrshort{gb_dt} model was only used to detect surface water during flooding, and no distinction was made between permanent water surfaces and flooded areas. However, since maps of permanent water are freely available (\cite{bonafiliaSen1Floods11GeoreferencedDataset2020}), this is not a limitation in practice, since permanent water regions such as lakes or rivers can simply be subtracted from the surface water mask to obtain a map of pure flooded areas. Accordingly, the proposed model is also suitable for the detection of pure flooded areas.
	
	Additionally, the success of such a simple pixel-wise model on optical data shows that, using the right feature space, little contextual information is needed to enable very accurate flood detection. Accordingly, methodologies using specialized architectures for precise segmentation of images, such as BASNet (\cite{qinBASNetBoundaryAwareSalient2019}) which is used by \cite{baiEnhancementDetectingPermanent2021}, are expected to be particularly successful provided optical data is available. However, this is also a potential challenge for future deep data fusion approaches using optical data, as the deep neural networks used may be inclined to rely solely on the high predictive capability of optical data without using the other modalities effectively.
	
	Lastly, while the proposed cassical machine learning models are already very well suited as standalone flood mapping classifiers, further interesting directions for future work emerge from the interaction with Deep Learning. For all algorithms it could be observed that the performance of the respective classifier strongly depends on the feature space used. ML models based on multispectral data outperform those based on only RGB or \acrshort{sar} data by a multiple. However, these feature spaces exhibit useful properties such as higher revisit-times or independence from weather conditions (\cite{shenInundationExtentMapping2019, mateo-garciaGlobalFloodMapping2021}) making them highly desirable flood mapping feature spaces. Since prior work has shown that deep learning models based on larger datasets with weak labels can outperform those based on smaller datasets, albeit higher quality labels (\cite{katiyarNearRealTimeFloodMapping2021, bonafiliaSen1Floods11GeoreferencedDataset2020}), further research could therefore analyze the potential of the described ML algorithms as label-sources for deep learning methodologies trained in a global context based on these more challenging feature spaces. 
	
	We believe that a classifier, which outperforms the previous state-of-the-art results in terms of \acrsmean{iou}, \acrstotal{iou} and \acrlmean{acc} with values of $0.7031$, $0.8751$ and $0.9718$ compared to $0.5873$, $0.70$ and $0.9338$, will serve both as a highly accurate flood inundation mapping methodology and a label-source for the development of future deep learning based approaches.
	\newpage
	\section*{Acknowledgments}
	This project was partly funded through the ESA InCubed Programme (\url{https://incubed.esa.int/}) as part of the project AI4EO Solution Factory (\url{https://www.ai4eo-solution-factory.de/})
	\bibliographystyle{elsarticle-harv} 
	\bibliography{literature.bib}
	\appendix 
	\section{Further Hyperparameter Analysis} \label{sec:appendix:hyperparameters}
	\begin{figure*}
		\centering
		\begin{subfigure}{0.45\linewidth}
			\centering
			\includegraphics[width=0.8\linewidth]{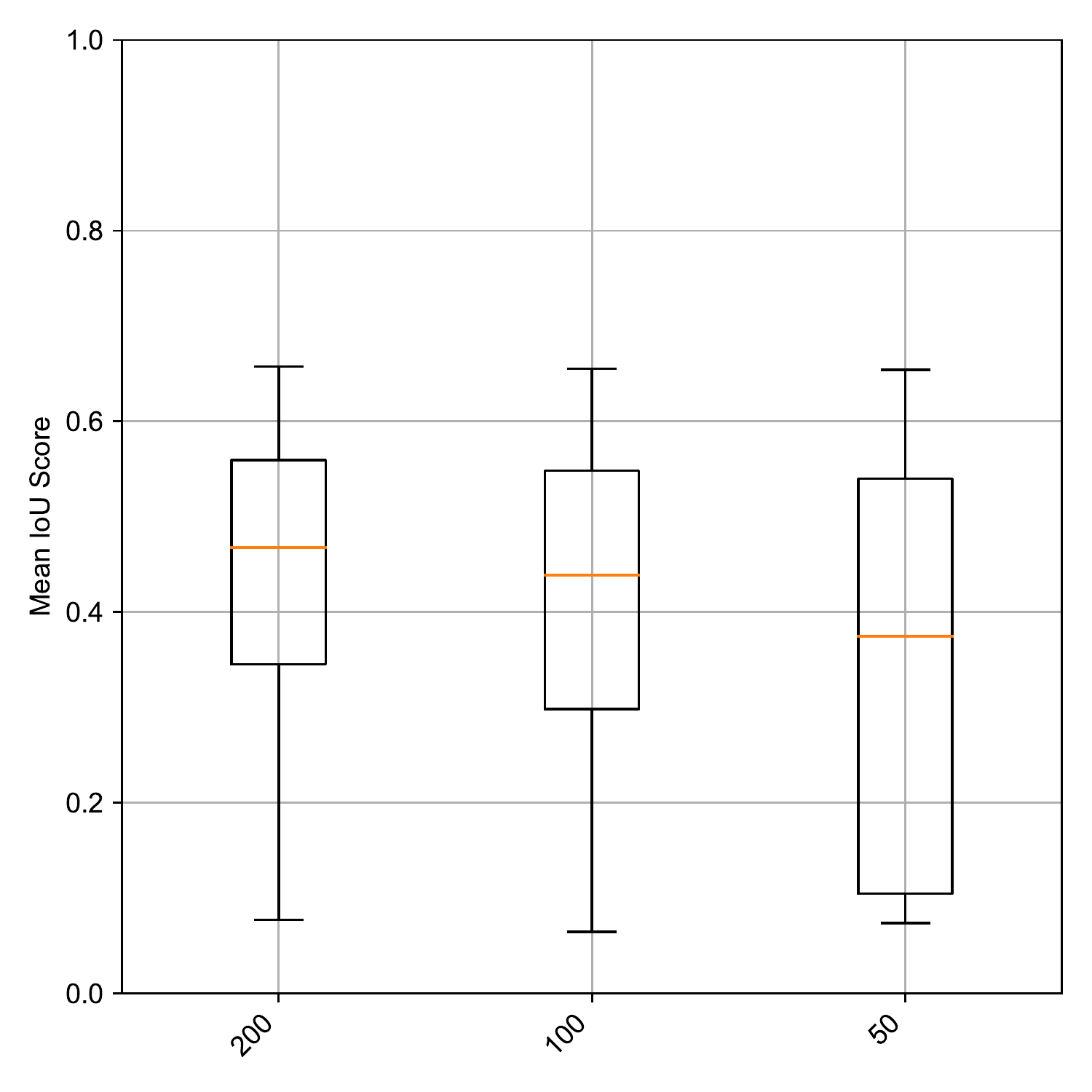} 
			\caption{Influence of the number of trees on \acrshort{gb_dt}}
			\label{fig:classifiers:models:n_estimators}
		\end{subfigure}
		\begin{subfigure}{0.45\linewidth}
			\centering
			\includegraphics[width=0.8\linewidth]{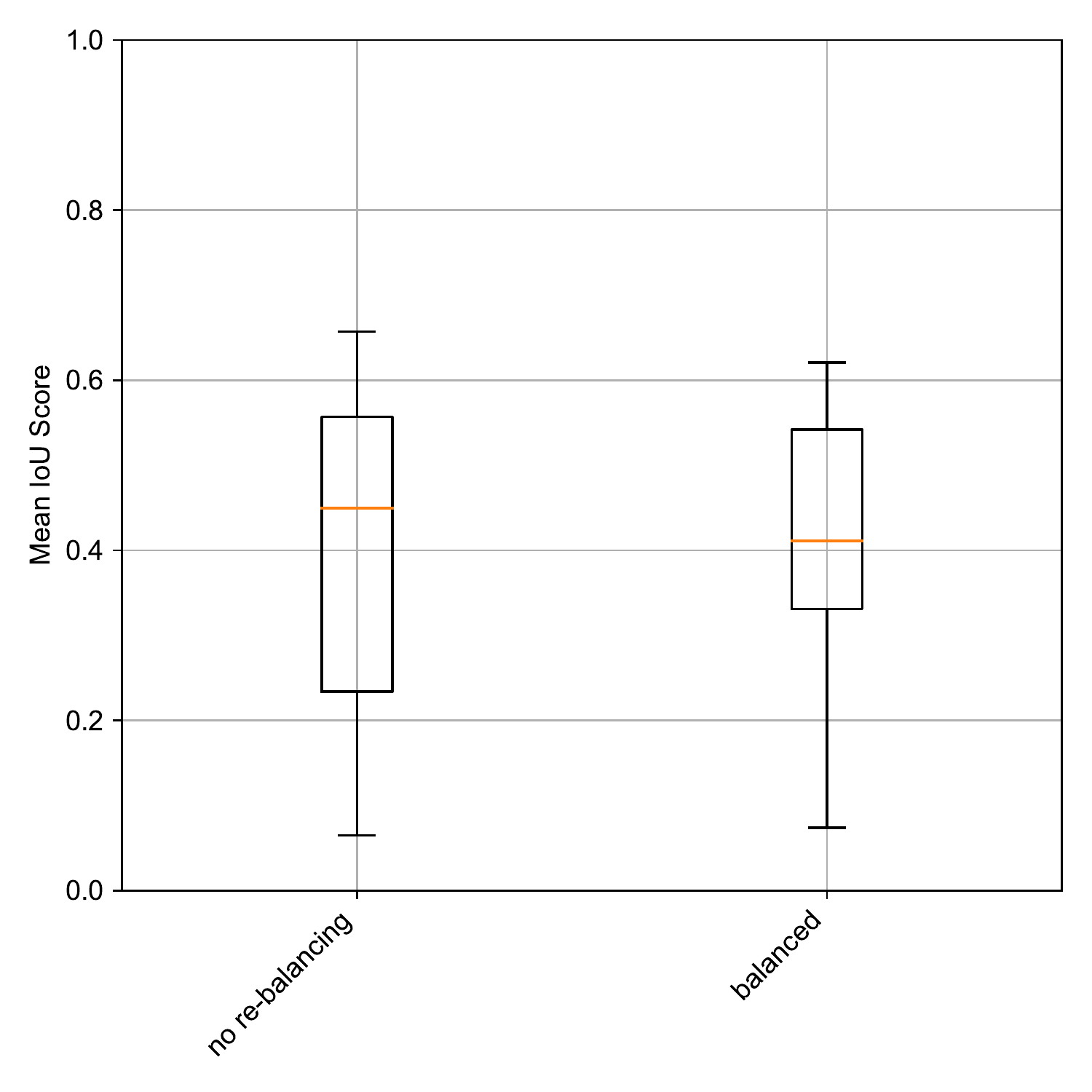} 
			\caption{Influence of class re-balancing on \acrshort{gb_dt}}
			\label{fig:classifiers:models:class_weight}
		\end{subfigure}
		\caption{Influence of tested parameters on \acrshort{gb_dt} in terms of \acrsmean{iou} on the Sen1Floods11 validation set. The x-Axis is ordered by the respective max-values, with whiskers depicting the corresponding min-max ranges.}
		\label{fig:appendix:additional_general_parameters}
	\end{figure*}
	\begin{figure*}
		\centering
		\begin{subfigure}{0.45\linewidth}
			\centering
			\includegraphics[width=\linewidth]{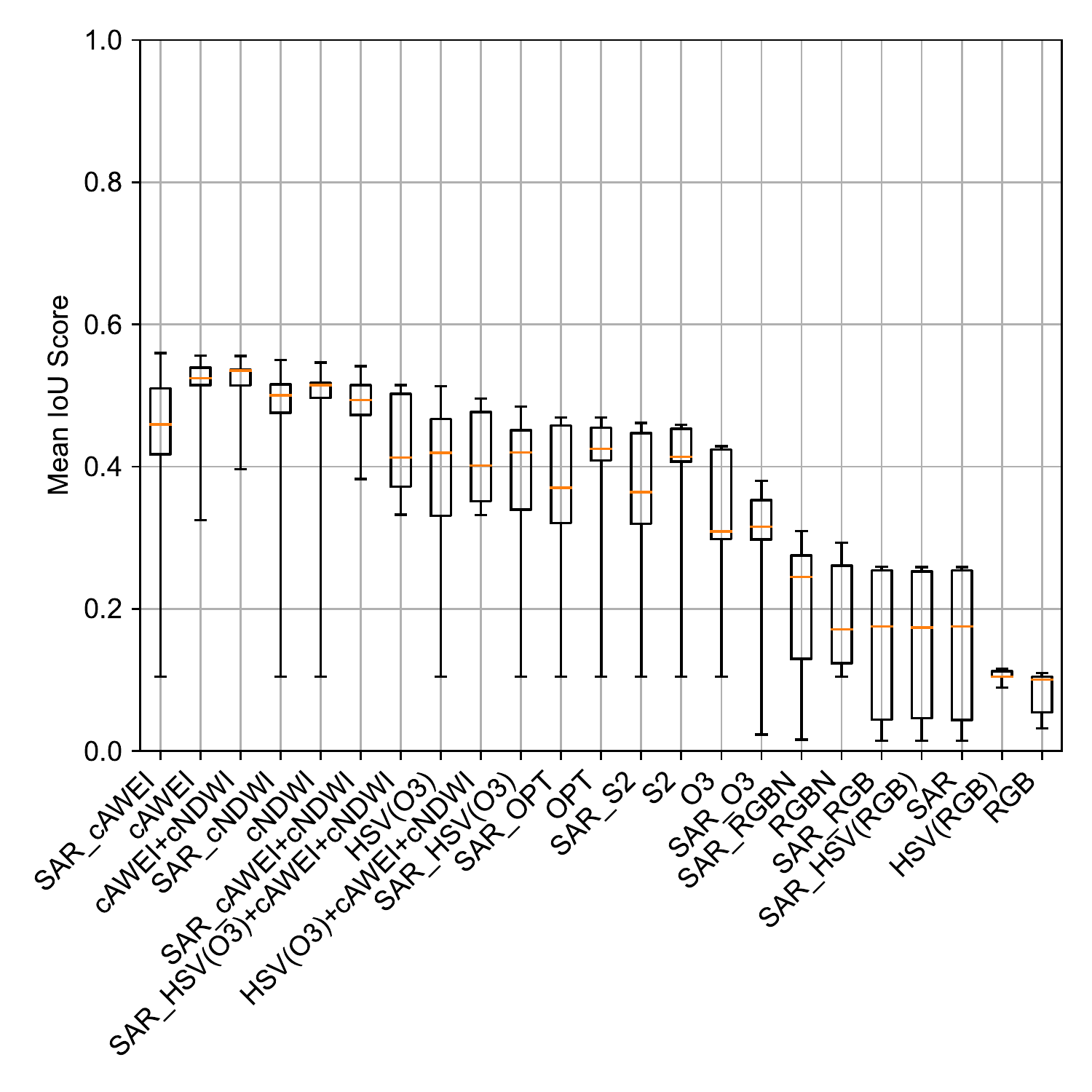} 
			\caption{Influence of the choice of feature space on the linear \acrshort{sgd} model.}
		\end{subfigure}
		\begin{subfigure}{0.45\linewidth}
			\centering
			\includegraphics[width=\linewidth]{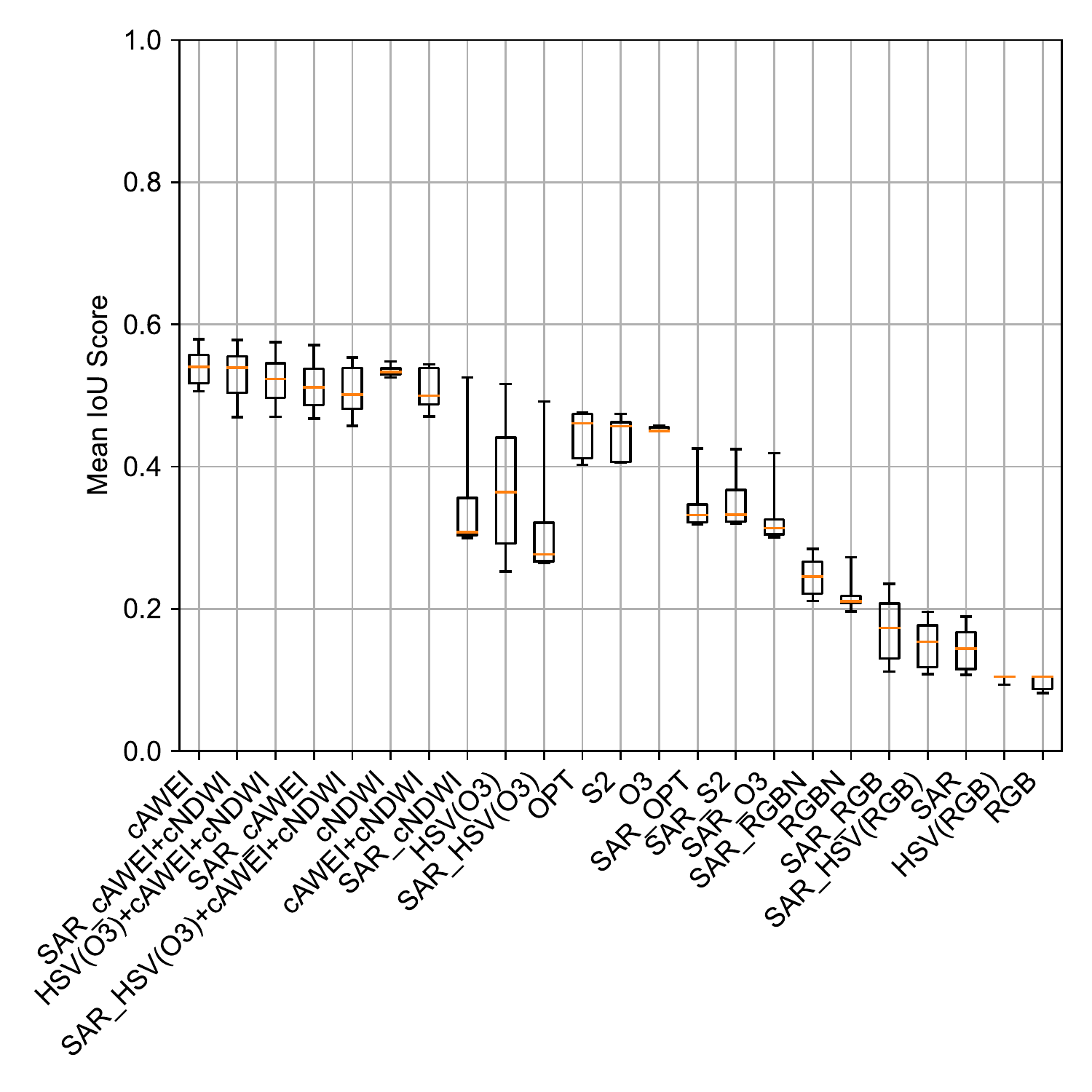} 
			\caption{Influence of the choice of feature space on \acrfull{lda}.}
		\end{subfigure}\\
		\begin{subfigure}{0.45\linewidth}
			\centering
			\includegraphics[width=\linewidth]{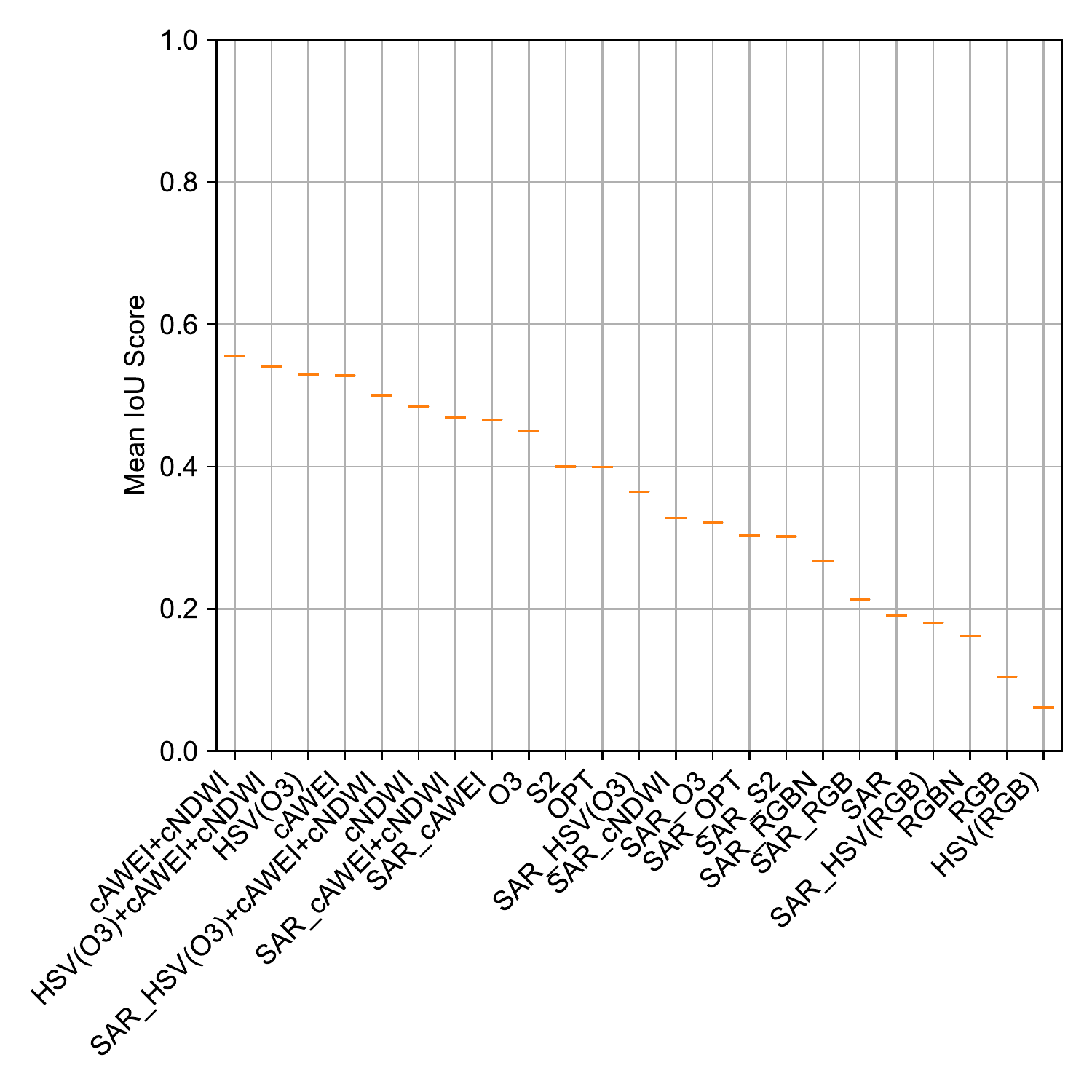} 
			\caption{Influence of the choice of feature space on the Naive Bayes classifier.}
		\end{subfigure}
		\begin{subfigure}{0.45\linewidth}
			\centering
			\includegraphics[width=\linewidth]{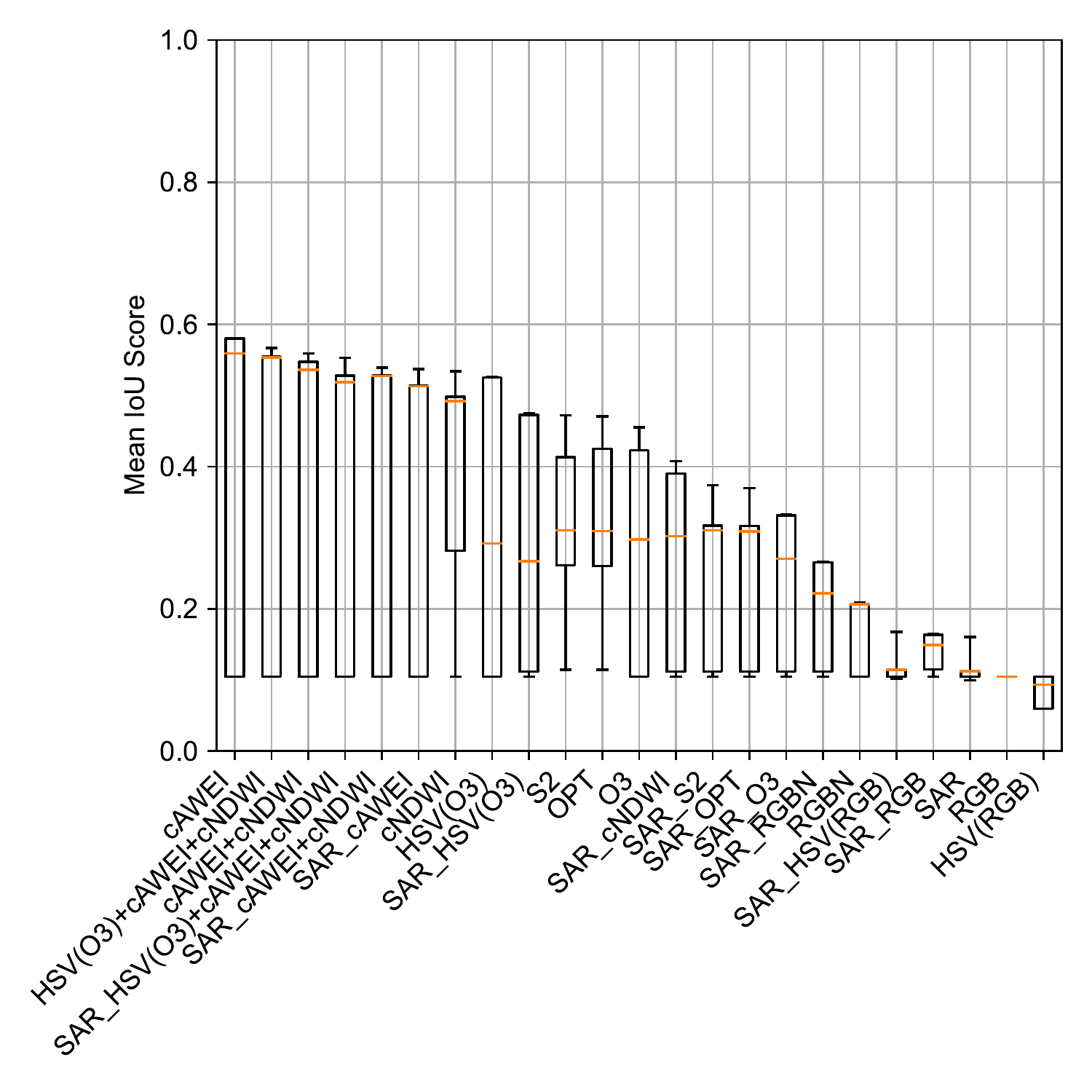} 
			\caption{Influence of the choice of feature space on \acrfull{qda}.}
		\end{subfigure}
		\caption{Influence of the choice of feature space on the tested classical machine learning models besides \acrshort{gb_dt} in terms of \acrsmean{iou} on the Sen1Floods11 validation set. Similar to Figure \ref{fig:classifiers:models:feature_spaces}, the x-Axis is ordered by the respective max-values, with whiskers depicting the corresponding min-max ranges.}
		\label{fig:appendix:additional_feature_spaces}
	\end{figure*}
	More details to the results of our hyperparameter search for \acrshort{gb_dt} can be found in Figure \ref{fig:appendix:additional_general_parameters}. Additionally, we observed that for all considered machine learning models, the feature space forms the most important parameter. This is depicted in Figure \ref{fig:appendix:additional_feature_spaces}.
	\section{Additional Quantitative Results} \label{sec:appendix:quantitative}
	\begin{table*}
		\centering
		\caption[Results reported on the Sen1Floods11 dataset]{Extended performance comparison of the proposed classifiers using both data modalities, therefore Sentinel-1 (\acrshort{sar}) data and Sentinel-2 (optical) data, with related work.  Metric values from related work on the Sen1Floods11 dataset are taken from the individual papers and, if available, ranked by \acrstotal{iou} and otherwise \acrsmean{iou} within the evaluation split to compare them with the proposed classical machine learning approaches. The independently identically distributed (IID) split hereby refers to the regular Sen1Floods11 test set, whereas the domain shifted split refers to Bolivia test split. For each method, the used feature space as well as other parameter choices (such as hyperparameters or whether additional data has been used) are provided for a more in-depth comparison. Note that the reported standard deviation corresponds to the variation across the test set images, and not to the variation across individual runs with different seeds.}
		\resizebox{\textwidth}{!}{
			\begin{tabular}{cccccccccccc}
				\toprule %
				\multirowcell{2}{Test Split} & \multirowcell{2}{Method} & \multirowcell{2}{Feature Space} & \multirowcell{2}{Mean \acrshort{iou}\\flooded (std)} &  \multirowcell{2}{Total \acrshort{iou}\\flooded} & \multirowcell{2}{Mean \acrlong{acc}\\(std)} & \multirowcell{2}{Total \acrlong{acc}} & \multirowcell{2}{Total \acrlong{pr}\\flooded} & \multirowcell{2}{Total \acrlong{re}\\flooded} & \multirowcell{2}{Total \acrlong{re}\\dry} & \multirowcell{2}{Total \acrlong{f1}\\flooded} & \multirowcell{2}{Parameter choices}\\ 
				& & & & & & & & & & & \\ \midrule
				
				\multirowcell{30}{\emph{IID}\\\emph{Split}} & \multirowcell{5}{\acrshort{gb_dt}} & \multirowcell{5}{SAR$\_$HSV(O3)+cAWEI+cNDWI} & \multirowcell{5}{$\mathbf{0.7031} (\pm 0.2984)$} & \multirowcell{5}{$\mathbf{0.8751}$} & \multirowcell{5}{$\mathbf{0.9718} (\pm 0.1176)$} & \multirowcell{5}{$\mathbf{0.9838}$} & \multirowcell{5}{$\mathbf{0.9577}$} & \multirowcell{5}{$0.9103$} & \multirowcell{5}{$\mathbf{0.9943}$} & \multirowcell{5}{$\mathbf{0.9334}$} & $200$ trees\\ 
				& & & & & & & & & & & up to $128$ leaves per tree\\
				& & & & & & & & & & & regularisation $\lambda=1$ \\ 
				& & & & & & & & & & & learning rate: $0.1$ \\ 
				& & & & & & & & & & & subsample size: $262144$ \\ \cmidrule{2-12}
				
				& \multirowcell{3}{Linear\\Model\\(\acrshort{sgd})} & \multirowcell{3}{SAR$\_$HSV(O3)} & \multirowcell{3}{$0.5701 (\pm 0.3305)$} & \multirowcell{3}{$0.7966$} & \multirowcell{3}{$0.9627 (\pm 0.1127)$} & \multirowcell{3}{$0.9724$} & \multirowcell{3}{$0.9172$} & \multirowcell{3}{$0.8569$} & \multirowcell{3}{$0.9889$} & \multirowcell{3}{$0.8860$} & logistic loss\\ 
				& & & & & & & & & & & no class rebalancing\\ 
				& & & & & & & & & & & regularisation $\alpha=0.0001$\\ \cmidrule{2-12}
				
				& \multirowcell{3}{Linear\\Discriminant\\Analysis} & \multirowcell{3}{SAR$\_$cAWEI+cNDWI} & \multirowcell{3}{$0.5631 (\pm 0.3262)$} & \multirowcell{3}{$0.7828$} & \multirowcell{3}{$0.9618 (\pm 0.0906)$} & \multirowcell{3}{$0.9685$} & \multirowcell{3}{$0.8540$} & \multirowcell{3}{$0.9048$} & \multirowcell{3}{$0.9776$} & \multirowcell{3}{$0.8780$} & \multirowcell{3}{shrinkage $\rho = 0.1$} \\ 
				& & & & & & & & & & & \\ 
				& & & & & & & & & & & \\ \cmidrule{2-12}
				
				& \multirowcell{3}{Quadratic\\Discriminant\\Analysis} & \multirowcell{3}{SAR$\_$HSV(O3)+cAWEI+cNDWI} & \multirowcell{3}{$0.5679 (\pm 0.3232)$} & \multirowcell{3}{$0.7695$} & \multirowcell{3}{$0.9631 (\pm 0.0654)$} & \multirowcell{3}{$0.9653$} & \multirowcell{3}{$0.8201$} & \multirowcell{3}{$0.9258$} & \multirowcell{3}{$0.9710$} & \multirowcell{3}{$0.8698$} & \multirowcell{3}{regularisation: $0$} \\ 
				& & & & & & & & & & & \\ 
				& & & & & & & & & & & \\ \cmidrule{2-12}
				
				& \multirowcell{2}{Naive\\Bayes} & \multirowcell{2}{SAR$\_$HSV(O3)+cAWEI+cNDWI} & \multirowcell{2}{$0.5187 (\pm 0.3168)$} & \multirowcell{2}{$0.7399$} & \multirowcell{2}{$0.9522 (\pm 0.0860)$} & \multirowcell{2}{$0.9586$} & \multirowcell{2}{$0.7784$} & \multirowcell{2}{$\mathbf{0.9378}$} & \multirowcell{2}{$0.9616$} & \multirowcell{2}{$0.8503$} & \multirowcell{2}{-}\\ 
				& & & & & & & & & & & \\ \cmidrule{2-12}
				
				& \multirowcell{2}{\cite{yadavAttentiveDualStream2022}} & \multirowcell{2}{SAR} & \multirowcell{2}{-} & \multirowcell{2}{$0.70$} & \multirowcell{2}{-} & \multirowcell{2}{-} & \multirowcell{2}{-} & \multirowcell{2}{-} & \multirowcell{2}{-} & \multirowcell{2}{0.83} & Attentive Dual Stream Siamese Network \\ 
				& & & & & & & & & & & Additional Pre-Flood \acrshort{sar} Data\\  \cmidrule{2-12}
				
				& \multirowcell{3}{\cite{jainMultimodalContrastiveLearning2022}} & \multirowcell{3}{SAR} & \multirowcell{3}{-} & \multirowcell{3}{$0.6871$} & \multirowcell{3}{-} & \multirowcell{3}{-} & \multirowcell{3}{-} & \multirowcell{3}{-} & \multirowcell{3}{-} & \multirowcell{3}{-} & DeepLabV3+ (\cite{chenEncoderDecoderAtrousSeparable2018})\\ 
				& & & & & & & & & & & Multimodal contrastive pretraining \\ 
				& & & & & & & & & & & 1,087,502 additional unlabeled images  \\ \cmidrule{2-12}
				
				& \multirowcell{3}{\cite{patelEvaluatingSelfSemiSupervised2021}} & \multirowcell{3}{SAR} & \multirowcell{3}{-} & \multirowcell{3}{$0.6692$} & \multirowcell{3}{-} & \multirowcell{3}{-} & \multirowcell{3}{-} & \multirowcell{3}{-} & \multirowcell{3}{-} & \multirowcell{3}{-} & DeepLabV3+ (\cite{chenEncoderDecoderAtrousSeparable2018})\\ 
				& & & & & & & & & & & SimCLR pretraining \\ 
				& & & & & & & & & & & 67k additional unlabeled images  \\ \cmidrule{2-12}
				
				& \multirowcell{3}{\cite{baiEnhancementDetectingPermanent2021}} & \multirowcell{3}{SAR$\_$S2} & \multirowcell{3}{$0.5873$} & \multirowcell{3}{$0.6452$} & \multirowcell{3}{$0.9338$} & \multirowcell{3}{-} & \multirowcell{3}{-} & \multirowcell{3}{$0.6881$} & \multirowcell{3}{$0.9884$} & \multirowcell{3}{-} & BASNet (\cite{qinBASNetBoundaryAwareSalient2019}) \\ 
				& & & & & & & & & & & \multirowcell{2}{Uses Sen1Floods11 weakly-labeled\\split for pretraining (4384 images)} \\ 
				& & & & & & & & & & &\\ \cmidrule{2-12}
				
				& \multirowcell{3}{\cite{katiyarNearRealTimeFloodMapping2021}} & \multirowcell{3}{SAR} & \multirowcell{3}{$0.494$} & \multirowcell{3}{-} & \multirowcell{3}{-} & \multirowcell{3}{-} & \multirowcell{3}{-} & \multirowcell{3}{$0.7958$} & \multirowcell{3}{$0.9796$} & \multirowcell{3}{-} & \multirowcell{2}{U-Net (\cite{UNet})\\Uses Sen1Floods11 weakly-labeled\\split for pretraining (4384 images)} \\ 
				& & & & & & & & & & &  \\ 
				& & & & & & & & & & &\\ \cmidrule{2-12}
				
				& \multirowcell{3}{Sentinel-2 Weak\\\cite{bonafiliaSen1Floods11GeoreferencedDataset2020}} & \multirowcell{3}{SAR} & \multirowcell{3}{$0.4084$} & \multirowcell{3}{-} & \multirowcell{3}{$0.9384$} & \multirowcell{3}{-} & \multirowcell{3}{-} & \multirowcell{3}{$0.7518$} & \multirowcell{3}{$0.9222$} & \multirowcell{3}{-} & Resnet-50 (\cite{heDeepResidualLearning2016})  \\ 
				& & & & & & & & & & & \multirowcell{2}{Uses Sen1Floods11 weakly-labeled\\split for pretraining (4384 images)} \\ 
				& & & & & & & & & & &\\ \cmidrule{2-12}
				
				& \multirowcell{2}{Otsu-Reference\\ \cite{bonafiliaSen1Floods11GeoreferencedDataset2020}} & \multirowcell{2}{SAR} & \multirowcell{2}{$0.3591$} & \multirowcell{2}{-} & \multirowcell{2}{$0.9389$} & \multirowcell{2}{-} & \multirowcell{2}{-} & \multirowcell{2}{$0.8573$} & \multirowcell{2}{$0.9151$} & \multirowcell{2}{-} & \multirowcell{2}{VH-Threshold}\\ 
				& & & & & & & & & & & \\ \cmidrule{2-12}
				
				& \multirowcell{3}{Sentinel-1 Weak\\\cite{bonafiliaSen1Floods11GeoreferencedDataset2020}} & \multirowcell{3}{SAR} & \multirowcell{3}{$0.3092$} & \multirowcell{3}{-} & \multirowcell{3}{-} & \multirowcell{3}{-} & \multirowcell{3}{-} & \multirowcell{3}{$0.8876$} & \multirowcell{3}{$0.9003$} & \multirowcell{3}{-} & Resnet-50 (\cite{heDeepResidualLearning2016})  \\ 
				& & & & & & & & & & & \multirowcell{2}{Uses Sen1Floods11 weakly-labeled\\split for pretraining (4384 images)} \\ 
				& & & & & & & & & & &\\  \midrule
				
				\multirowcell{30}{\emph{Domain}\\\emph{Shifted}\\\emph{Split}} & \multirowcell{5}{\acrshort{gb_dt}} & \multirowcell{5}{SAR$\_$HSV(O3)+cAWEI+cNDWI} & \multirowcell{5}{$\mathbf{0.6070} (\pm 0.3542)$} & \multirowcell{5}{$\mathbf{0.8357}$} & \multirowcell{5}{$\mathbf{0.9705} (\pm 0.0293)$} & \multirowcell{5}{$\mathbf{0.9730}$} & \multirowcell{5}{$\mathbf{0.9584}$} & \multirowcell{5}{$0.8671$} & \multirowcell{5}{$\mathbf{0.9929}$} & \multirowcell{5}{$\mathbf{0.9105}$} & $200$ trees \\ 
				& & & & & & & & & & & up to $128$ leaves per tree\\ 
				& & & & & & & & & & & regularisation $\lambda=1$\\ 
				& & & & & & & & & & & learning rate: $0.1$\\ 
				& & & & & & & & & & & subsample size: $262144$\\ \cmidrule{2-12}
				
				& \multirowcell{3}{Linear\\Model\\(\acrshort{sgd})} & \multirowcell{3}{SAR$\_$HSV(O3)} & \multirowcell{3}{$0.5599 (\pm 0.3231)$} & \multirowcell{3}{$0.7900$} & \multirowcell{3}{$0.9581 (\pm 0.0382)$} & \multirowcell{3}{$0.9615$} & \multirowcell{3}{$0.8768$} & \multirowcell{3}{$0.8925$} & \multirowcell{3}{$0.9746$} & \multirowcell{3}{$0.8820$} & logistic loss\\ 
				& & & & & & & & & & & regularisation $\alpha=0.0001$\\ 
				& & & & & & & & & & & no class rebalancing\\ \cmidrule{2-12}
				
				& \multirowcell{3}{\cite{baiEnhancementDetectingPermanent2021}} & \multirowcell{3}{SAR$\_$S2} & \multirowcell{3}{$0.5407$} & \multirowcell{3}{$0.7890$} & \multirowcell{3}{$0.9579$} & \multirowcell{3}{-} & \multirowcell{3}{-} & \multirowcell{3}{$0.9234$} & \multirowcell{3}{$0.9679$} & \multirowcell{3}{-} & BASNet (\cite{qinBASNetBoundaryAwareSalient2019}) \\ 
				& & & & & & & & & & & \multirowcell{2}{Uses Sen1Floods11 weakly-labeled\\split for pretraining (4384 images)} \\ 
				& & & & & & & & & & &\\ \cmidrule{2-12}
				
				& \multirowcell{3}{\cite{patelEvaluatingSelfSemiSupervised2021}} & \multirowcell{3}{SAR} & \multirowcell{3}{-} & \multirowcell{3}{$0.7156$} & \multirowcell{3}{-} & \multirowcell{3}{-} & \multirowcell{3}{-} & \multirowcell{3}{-} & \multirowcell{3}{-} & \multirowcell{3}{-} & DeepLabV3+ (\cite{chenEncoderDecoderAtrousSeparable2018})\\ 
				& & & & & & & & & & & FixMatch pretraining \\ 
				& & & & & & & & & & & 67k additional unlabeled images  \\ \cmidrule{2-12}
				
				& \multirowcell{3}{Linear\\Discriminant\\Analysis} & \multirowcell{3}{SAR$\_$cAWEI+cNDWI}  & \multirowcell{3}{$0.4881 (\pm 0.2924)$} & \multirowcell{3}{$0.6776$} & \multirowcell{3}{$0.9196 (\pm 0.0612)$} & \multirowcell{3}{$0.9241$} & \multirowcell{3}{$0.7027$} & \multirowcell{3}{$0.9557$} & \multirowcell{3}{$0.9181$} & \multirowcell{3}{$0.8050$} & \multirowcell{3}{shrinkage $\rho = 0.1$} \\ 
				& & & & & & & & & & & \\ 
				& & & & & & & & & & & \\ \cmidrule{2-12}
				
				& \multirowcell{3}{Quadratic\\Discriminant\\Analysis} & \multirowcell{3}{SAR$\_$HSV(O3)+cAWEI+cNDWI} & \multirowcell{3}{$0.4485 (\pm 0.2880)$} & \multirowcell{3}{$0.6554$} & \multirowcell{3}{$0.9163 (\pm 0.0680)$} & \multirowcell{3}{$0.9205$} & \multirowcell{3}{$0.6789$} & \multirowcell{3}{$0.9500$} & \multirowcell{3}{$0.9150$} & \multirowcell{3}{$0.7916$} & \multirowcell{3}{regularisation: $0$} \\ 
				& & & & & & & & & & & \\ 
				& & & & & & & & & & & \\ \cmidrule{2-12}
				
				& \multirowcell{2}{Naive\\Bayes} & \multirowcell{2}{SAR$\_$HSV(O3)+cAWEI+cNDWI} & \multirowcell{2}{$0.4402 (\pm 0.2889)$} & \multirowcell{2}{$0.6160$} & \multirowcell{2}{$0.8984 (\pm 0.0776)$} & \multirowcell{2}{$0.9033$} & \multirowcell{2}{$0.6268$} & \multirowcell{2}{$\mathbf{0.9728}$} & \multirowcell{2}{$0.8902$} & \multirowcell{2}{$0.7620$} & \multirowcell{2}{-}\\ 
				& & & & & & & & & & & \\ \cmidrule{2-12}
				
				& \multirowcell{3}{Sentinel-1 Weak\\\cite{bonafiliaSen1Floods11GeoreferencedDataset2020}} & \multirowcell{3}{SAR} & \multirowcell{3}{$0.3871$} & \multirowcell{3}{-} & \multirowcell{3}{$0.9384$} & \multirowcell{3}{-} & \multirowcell{3}{-} & \multirowcell{3}{$0.7213$} & \multirowcell{3}{$0.9305$} & \multirowcell{3}{-} & Resnet-50 (\cite{heDeepResidualLearning2016})  \\ 
				& & & & & & & & & & & \multirowcell{2}{Uses Sen1Floods11 weakly-labeled\\split for pretraining (4384 images)} \\ 
				& & & & & & & & & & &\\ \cmidrule{2-12}
				
				& \multirowcell{2}{Otsu-Reference\\ \cite{bonafiliaSen1Floods11GeoreferencedDataset2020}} & \multirowcell{2}{SAR} &  \multirowcell{2}{$0.3862$} & \multirowcell{2}{-} & \multirowcell{2}{-} & \multirowcell{2}{-} & \multirowcell{2}{-} & \multirowcell{2}{$0.7520$} & \multirowcell{2}{$0.9475$} & \multirowcell{2}{-} & \multirowcell{2}{VH-Threshold}\\ 
				& & & & & & & & & & \\ \cmidrule{2-12}
				
				& \multirowcell{3}{Sentinel-2 Weak\\\cite{bonafiliaSen1Floods11GeoreferencedDataset2020}} & \multirowcell{3}{SAR} & \multirowcell{3}{$0.3160$} & \multirowcell{3}{-} & \multirowcell{3}{-} & \multirowcell{3}{-} & \multirowcell{3}{-} & \multirowcell{3}{$0.8425$} & \multirowcell{3}{$0.9586$} & \multirowcell{3}{-} & Resnet-50 (\cite{heDeepResidualLearning2016})  \\ 
				& & & & & & & & & & & \multirowcell{2}{Uses Sen1Floods11 weakly-labeled\\split for pretraining (4384 images)} \\ 
				& & & & & & & & & & &\\ \bottomrule
			\end{tabular}
		}
		\label{tab:classifier_comparison_combined_full}
	\end{table*} 
	\begin{table*}
		\centering
		\caption[Results of the tested classifiers on Sentinel-2 (optical) data]{Performance comparison of the best classifiers for each model using only Sentinel-2 (optical) data, in the same format as Table \ref{tab:classifier_comparison_combined_full}.}
		\resizebox{\textwidth}{!}{
			\begin{tabular}{cccccccccccc} \toprule
				\multirowcell{2}{Test Split} & \multirowcell{2}{Method} & \multirowcell{2}{Feature Space} & \multirowcell{2}{Mean \acrshort{iou}\\flooded (std)} &  \multirowcell{2}{Total \acrshort{iou}\\flooded} & \multirowcell{2}{Mean \acrlong{acc}\\(std)} & \multirowcell{2}{Total \acrlong{acc}} & \multirowcell{2}{Total \acrlong{pr}\\flooded} & \multirowcell{2}{Total \acrlong{re}\\flooded} & \multirowcell{2}{Total \acrlong{re}\\dry} & \multirowcell{2}{Total \acrlong{f1}\\flooded} & \multirowcell{2}{Parameter choices}\\ 
				& & & & & & & & & & & \\ \midrule
				
				\multirowcell{16}{\emph{IID}\\\emph{Split}} & \multirowcell{5}{\acrshort{gb_dt}} & \multirowcell{5}{HSV(O3)+cAWEI+cNDWI} & \multirowcell{5}{$\mathbf{0.7010} (\pm 0.3001)$} & \multirowcell{5}{$\mathbf{0.8767}$} & \multirowcell{5}{$\mathbf{0.9719} (\pm 0.1170)$} & \multirowcell{5}{$\mathbf{0.9839}$} & \multirowcell{5}{$\mathbf{0.9568}$} & \multirowcell{5}{$0.9129$} & \multirowcell{5}{$\mathbf{0.9941}$} & \multirowcell{5}{$\mathbf{0.9343}$} & $200$ trees\\ 
				& & & & & & & & & & & up to $64$ leaves per tree\\ 
				& & & & & & & & & & & regularisation $\lambda=1$\\ 
				& & & & & & & & & & & learning rate: $0.1$\\ 
				& & & & & & & & & & & subsample size: $1048576$\\ \cmidrule{2-12}
				
				& \multirowcell{3}{Linear\\Discriminant\\Analysis} & \multirowcell{3}{cAWEI} & \multirowcell{3}{$0.5928 (\pm 0.3199)$} & \multirowcell{3}{$0.8140$} & \multirowcell{3}{$0.9621 (\pm 0.1141)$} & \multirowcell{3}{$0.9739$} & \multirowcell{3}{$0.8819$} & \multirowcell{3}{$0.9136$} & \multirowcell{3}{$0.9825$} & \multirowcell{3}{$0.8975$}& \multirowcell{3}{shrinkage $\rho = 0$}\\ 
				& & & & & & & & & &\\ 
				& & & & & & & & & &\\ \cmidrule{2-12}
				
				& \multirowcell{3}{Quadratic\\Discriminant\\Analysis} & \multirowcell{3}{cAWEI} & \multirowcell{3}{$0.5880 (\pm 0.3211)$} & \multirowcell{3}{$0.8139$} & \multirowcell{3}{$0.9622 (\pm 0.1150)$} & \multirowcell{3}{$0.9740$} & \multirowcell{3}{$0.8849$} & \multirowcell{3}{$0.9104$} & \multirowcell{3}{$0.9831$} & \multirowcell{3}{$0.8974$}& \multirowcell{3}{regularisation: $0.001$} \\ 
				& & & & & & & & & & &\\ 
				& & & & & & & & & & &\\ \cmidrule{2-12}
				
				& \multirowcell{3}{Linear\\Model\\(\acrshort{sgd})} & \multirowcell{3}{HSV(O3)}& \multirowcell{3}{$0.5742 (\pm 0.3336)$} & \multirowcell{3}{$0.7908$} & \multirowcell{3}{$0.9612 (\pm 0.1160)$} & \multirowcell{3}{$0.9712$} & \multirowcell{3}{$0.9104$} & \multirowcell{3}{$0.8547$} & \multirowcell{3}{$0.9879$} & \multirowcell{3}{$0.8816$} & logistic loss \\ 
				& & & & & & & & & & & regularisation $\alpha=0.0001$\\ 
				& & & & & & & & & & & no class rebalancing\\ \cmidrule{2-12}
				
				& \multirowcell{2}{Naive\\Bayes} & \multirowcell{2}{cAWEI+cNDWI}  & \multirowcell{2}{$0.5511 (\pm 0.3177)$} & \multirowcell{2}{$0.7343$} & \multirowcell{2}{$0.9551 (\pm 0.0636)$} & \multirowcell{2}{$0.9574$} & \multirowcell{2}{$0.7692$} & \multirowcell{2}{$\mathbf{0.9418}$} & \multirowcell{2}{$0.9596$} & \multirowcell{2}{$0.8468$}&\multirowcell{2}{-}\\ 
				& & & & & & & & & & &\\  \midrule
				
				\multirowcell{16}{\emph{Domain}\\\emph{Shifted}\\\emph{Split}} & \multirowcell{5}{\acrshort{gb_dt}} & \multirowcell{5}{HSV(O3)+cAWEI+cNDWI} & \multirowcell{5}{$\mathbf{0.6006} (\pm 0.3537)$} & \multirowcell{5}{$\mathbf{0.8295}$} & \multirowcell{5}{$0.9692 (\pm 0.0312)$} & \multirowcell{5}{$0.9719$} & \multirowcell{5}{$\mathbf{0.9571}$} & \multirowcell{5}{$0.8616$} & \multirowcell{5}{$\mathbf{0.9927}$} & \multirowcell{5}{$\mathbf{0.9068}$} & $200$ trees\\ 
				& & & & & & & & & & & up to $64$ leaves per tree\\ 
				& & & & & & & & & & & regularisation $\lambda=1$\\ 
				& & & & & & & & & & & learning rate: $0.1$\\ 
				& & & & & & & & & & & subsample size: $1048576$\\ \cmidrule{2-12}
				
				& \multirowcell{3}{Quadratic\\Discriminant\\Analysis} & \multirowcell{3}{cAWEI} & \multirowcell{3}{$0.5804 (\pm 0.3350)$} & \multirowcell{3}{$0.7996$} & \multirowcell{3}{$\mathbf{0.9739} (\pm 0.0355)$} & \multirowcell{3}{$\mathbf{0.9748}$} & \multirowcell{3}{$0.8676$} & \multirowcell{3}{$0.9107$} & \multirowcell{3}{$0.9828$} & \multirowcell{3}{$0.8886$}& \multirowcell{3}{regularisation: $0.001$} \\ 
				& & & & & & & & & &\\ 
				& & & & & & & & & &\\ \cmidrule{2-12}
				
				& \multirowcell{3}{Linear\\Model\\(\acrshort{sgd})} & \multirowcell{3}{HSV(O3)}& \multirowcell{3}{$0.5458 (\pm 0.3135)$} & \multirowcell{3}{$0.7796$} & \multirowcell{3}{$0.9551 (\pm 0.0379)$} & \multirowcell{3}{$0.9585$} & \multirowcell{3}{$0.8649$} & \multirowcell{3}{$0.8936$} & \multirowcell{3}{$0.9707$} & \multirowcell{3}{$0.8750$} & logistic loss \\ 
				& & & & & & & & & & & regularisation $\alpha=0.0001$\\ 
				& & & & & & & & & & & no class rebalancing\\ \cmidrule{2-12}
				
				& \multirowcell{3}{Linear\\Discriminant\\Analysis} & \multirowcell{3}{cAWEI} & \multirowcell{3}{$0.5394 (\pm 0.2922)$} & \multirowcell{3}{$0.7451$} & \multirowcell{3}{$0.9436 (\pm 0.0533)$} & \multirowcell{3}{$0.9482$} & \multirowcell{3}{$0.7730$} & \multirowcell{3}{$0.9537$} & \multirowcell{3}{$0.9472$} & \multirowcell{3}{$0.8539$}& \multirowcell{3}{shrinkage $\rho = 0$}\\ 
				& & & & & & & & & & &\\ 
				& & & & & & & & & & &\\ \cmidrule{2-12}
				
				& \multirowcell{2}{Naive\\Bayes} & \multirowcell{2}{cAWEI+cNDWI}  & \multirowcell{2}{$0.4410 (\pm 0.2877)$} & \multirowcell{2}{$0.6290$} & \multirowcell{2}{$0.9039 (\pm 0.0741)$} & \multirowcell{2}{$0.9089$} & \multirowcell{2}{$0.6400$} & \multirowcell{2}{$\mathbf{0.9735}$} & \multirowcell{2}{$0.8968$} & \multirowcell{2}{$0.7723$} & \multirowcell{2}{-}\\ 
				& & & & & & & & & & &\\ \bottomrule
			\end{tabular}
		}
		\label{tab:classifier_comparison_optical}
	\end{table*} 
	\begin{table*}
		\centering
		\caption[Results of the tested classifiers on Sentinel-1 (\acrshort{sar}) data]{Performance comparison of the best classifiers for each model using only Sentinel-1 (\acrshort{sar}) data, in the same format as Table \ref{tab:classifier_comparison_combined_full}.}
		\resizebox{\textwidth}{!}{
			\begin{tabular}{cccccccccccc}\toprule
				\multirowcell{2}{Test Split} & \multirowcell{2}{Method} & \multirowcell{2}{Feature Space} & \multirowcell{2}{Mean \acrshort{iou}\\flooded (std)} &  \multirowcell{2}{Total \acrshort{iou}\\flooded} & \multirowcell{2}{Mean \acrlong{acc}\\(std)} & \multirowcell{2}{Total \acrlong{acc}} & \multirowcell{2}{Total \acrlong{pr}\\flooded} & \multirowcell{2}{Total \acrlong{re}\\flooded} & \multirowcell{2}{Total \acrlong{re}\\dry} & \multirowcell{2}{Total \acrlong{f1}\\flooded} & \multirowcell{2}{Parameter choices}\\ 
				& & & & & & & & & & & \\ \midrule
				
				\multirowcell{16}{\emph{IID}\\\emph{Split}} & \multirowcell{3}{Linear\\Model\\(\acrshort{sgd})} & \multirowcell{3}{SAR} & \multirowcell{3}{$0.2597 (\pm 0.2834)$} & \multirowcell{3}{$\mathbf{0.4968}$} & \multirowcell{3}{$0.8964 (\pm 0.1299)$} & \multirowcell{3}{$0.9063$} & \multirowcell{3}{$0.6022$} & \multirowcell{3}{$\mathbf{0.7396}$} & \multirowcell{3}{$0.9302$} & \multirowcell{3}{$\mathbf{0.6639}$} & hinge loss\\ 
				& & & & & & & & & & & regularisation $\alpha=0.1$\\ 
				& & & & & & & & & & & class rebalancing\\  \cmidrule{2-12}
				
				& \multirowcell{5}{\acrshort{gb_dt}} & \multirowcell{5}{SAR} & \multirowcell{5}{$\mathbf{0.2880} (\pm 0.3179)$} & \multirowcell{5}{$0.4713$} & \multirowcell{5}{$\mathbf{0.9207} (\pm 0.1644)$} & \multirowcell{5}{$\mathbf{0.9290}$} & \multirowcell{5}{$\mathbf{0.8722}$} & \multirowcell{5}{$0.5063$} & \multirowcell{5}{$\mathbf{0.9894}$} & \multirowcell{5}{$0.6407$} & $50$ trees\\ 
				& & & & & & & & & & & up to $2$ leaves per tree\\ 
				& & & & & & & & & & & regularisation $\lambda=0.01$\\
				& & & & & & & & & & & learning rate: $0.1$\\ 
				& & & & & & & & & & & subsample size: $262144$\\ \cmidrule{2-12}
				
				& \multirowcell{2}{Naive\\Bayes} & \multirowcell{2}{SAR} & \multirowcell{2}{$0.1852 (\pm 0.2594)$} & \multirowcell{2}{$0.4024$} & \multirowcell{2}{$0.9064 (\pm 0.1650)$} & \multirowcell{2}{$0.9142$} & \multirowcell{2}{$0.7571$} & \multirowcell{2}{$0.4621$} & \multirowcell{2}{$0.9788$} & \multirowcell{2}{$0.5739$} &\multirowcell{2}{-}\\ 
				& & & & & & & & & & &\\ \cmidrule{2-12}
				
				& \multirowcell{3}{Linear\\Discriminant\\Analysis} & \multirowcell{3}{SAR} & \multirowcell{3}{$0.1837 (\pm 0.2597)$} & \multirowcell{3}{$0.3960$} & \multirowcell{3}{$0.9057 (\pm 0.1684)$} & \multirowcell{3}{$0.9134$} & \multirowcell{3}{$0.7568$} & \multirowcell{3}{$0.4538$} & \multirowcell{3}{$0.9791$} & \multirowcell{3}{$0.5674$} & \multirowcell{3}{shrinkage $\rho = 1$}\\ 
				& & & & & & & & & & &\\
				& & & & & & & & & & &\\ \cmidrule{2-12}
				
				& \multirowcell{3}{Quadratic\\Discriminant\\Analysis} & \multirowcell{3}{SAR} & \multirowcell{3}{$0.1506 (\pm 0.2335)$} & \multirowcell{3}{$0.3345$} & \multirowcell{3}{$0.8986 (\pm 0.1745)$} & \multirowcell{3}{$0.9058$} & \multirowcell{3}{$0.7419$} & \multirowcell{3}{$0.3785$} & \multirowcell{3}{$0.9812$} & \multirowcell{3}{$0.5013$} & \multirowcell{3}{regularisation: $1$}\\
				& & & & & & & & & & & \\ 
				& & & & & & & & & & & \\ \midrule
				
				\multirowcell{16}{\emph{Domain}\\\emph{Shifted}\\\emph{Split}}  &\multirowcell{3}{Linear\\Model\\(\acrshort{sgd})} & \multirowcell{3}{SAR} & \multirowcell{3}{$\mathbf{0.4202} (\pm 0.3209)$} & \multirowcell{3}{$\mathbf{0.6552}$} & \multirowcell{3}{$0.9298 (\pm 0.0607)$} & \multirowcell{3}{$\mathbf{0.9328}$} & \multirowcell{3}{$0.7791$} & \multirowcell{3}{$\mathbf{0.8047}$} & \multirowcell{3}{$0.9570$} & \multirowcell{3}{$\mathbf{0.7917}$} & hinge loss\\ 
				& & & & & & & & & & & regularisation $\alpha=0.1$\\ 
				& & & & & & & & & & & class rebalancing\\  \cmidrule{2-12}
				
				& \multirowcell{5}{\acrshort{gb_dt}} & \multirowcell{5}{SAR}& \multirowcell{5}{$0.2842 (\pm 0.2578)$} & \multirowcell{5}{$0.4981$} & \multirowcell{5}{$0.9127 (\pm 0.0893)$} & \multirowcell{5}{$0.9182$} & \multirowcell{5}{$\mathbf{0.9487}$} & \multirowcell{5}{$0.5119$} & \multirowcell{5}{$\mathbf{0.9948}$} & \multirowcell{5}{$0.6650$} & $50$ trees\\ 
				& & & & & & & & & & & up to $2$ leaves per tree\\ 
				& & & & & & & & & & & regularisation $\lambda=0.01$\\
				& & & & & & & & & & & learning rate: $0.1$\\ 
				& & & & & & & & & & & subsample size: $262144$\\ \cmidrule{2-12}
				
				& \multirowcell{2}{Naive\\Bayes} & \multirowcell{2}{SAR} & \multirowcell{2}{$0.1628 (\pm 0.1527)$} & \multirowcell{2}{$0.2683$} & \multirowcell{2}{$0.8668 (\pm 0.1389)$} & \multirowcell{2}{$0.8731$} & \multirowcell{2}{$0.7592$} & \multirowcell{2}{$0.2933$} & \multirowcell{2}{$0.9825$} & \multirowcell{2}{$0.4231$} &\multirowcell{2}{-}\\ 
				& & & & & & & & & & &\\ \cmidrule{2-12}
				
				& \multirowcell{3}{Linear\\Discriminant\\Analysis} & \multirowcell{3}{SAR} & \multirowcell{3}{$0.1571 (\pm 0.1476)$} & \multirowcell{3}{$0.2531$} & \multirowcell{3}{$0.8644 (\pm 0.1434)$} & \multirowcell{3}{$0.8706$} & \multirowcell{3}{$0.7490$} & \multirowcell{3}{$0.2766$} & \multirowcell{3}{$0.9825$} & \multirowcell{3}{$0.4040$} & \multirowcell{3}{shrinkage $\rho = 1$}\\ 
				& & & & & & & & & & &\\
				& & & & & & & & & & &\\ \cmidrule{2-12}
				
				& \multirowcell{3}{Quadratic\\Discriminant\\Analysis} & \multirowcell{3}{SAR} & \multirowcell{3}{$0.0665 (\pm 0.0628)$} & \multirowcell{3}{$0.0957$} & \multirowcell{3}{$0.8354 (\pm 0.1853)$} & \multirowcell{3}{$0.8441$} & \multirowcell{3}{$0.5441$} & \multirowcell{3}{$0.1041$} & \multirowcell{3}{$0.9836$} & \multirowcell{3}{$0.1747$} & \multirowcell{3}{regularisation: $1$}\\
				& & & & & & & & & & & \\ 
				& & & & & & & & & & & \\  \bottomrule
				
			\end{tabular}
		}
		\label{tab:classifier_comparison_sar}
	\end{table*} 
	To provide a full quantitative analysis of the tested methodologies, Tables \ref{tab:classifier_comparison_combined_full}-\ref{tab:classifier_comparison_sar} extend Table \ref{tab:classifier_comparison_combined}. We depict the results for all tested methodologies on their best performing feature space in Table \ref{tab:classifier_comparison_combined_full} alongside all prior work that, to the best of our knowledge, has so far been carried out on the Sen1Floods11 dataset. Tables \ref{tab:classifier_comparison_optical} and \ref{tab:classifier_comparison_sar} additionally show the results of the proposed machine learning approaches using only optical and \acrshort{sar} data respectively.
\end{document}